\documentclass[11pt]{article}

\usepackage[preprint]{acl}
\usepackage{enumitem}

\usepackage{times}
\usepackage{latexsym}
\usepackage{amsmath}
\usepackage{wrapfig}
\usepackage{booktabs}
\usepackage{array}
\usepackage{makecell}
\usepackage[section]{placeins}
\usepackage[T1]{fontenc}

\usepackage[utf8]{inputenc}
\usepackage{newunicodechar}
\newunicodechar{≈}{\approx}

\usepackage{microtype}
\usepackage{listings}
\usepackage{inconsolata}
\usepackage{graphicx}
\usepackage{tikz}
\usepackage{tcolorbox}
\usepackage{varwidth}

\usepackage{caption}

\usetikzlibrary{positioning, shapes.geometric, arrows.meta, calc}

\definecolor{playerred}{RGB}{220, 50, 50}
\definecolor{playerblue}{RGB}{50, 120, 220}
\definecolor{lightred}{RGB}{255, 230, 230}
\definecolor{lightblue}{RGB}{230, 240, 255}

\title{The Language of Bargaining: Linguistic Effects in LLM Negotiations}

\author{
Stuti Sinha, Himanshu Kumar, Aryan Raju Mandapati, \\
{\bf Rakshit Sakhuja, Dhruv Kumar} \\
BITS Pilani \\
\texttt{\{f20220180, f20220557, f20220158, f20220471\}@pilani.bits-pilani.ac.in} \\
\texttt{dhruv.kumar@pilani.bits-pilani.ac.in}
}

\begin{document}
\maketitle

\begin{abstract}
Negotiation is a core component of social intelligence, requiring agents to balance strategic reasoning, cooperation, and social norms. Recent work shows that LLMs can engage in multi-turn negotiation, yet nearly all evaluations occur exclusively in English. We systematically isolate language effects across English and three Indic framings (Hindi, Punjabi, Gujarati) by holding game rules, model parameters, and incentives constant for multi-agent simulations for Ultimatum, Buy-Sell, and Resource Exchange games. Our results indicate that language choice is correlated to negotiation outcomes and that this correlation persists across model architectures. Crucially, effects are task-contingent: Indic languages reduce stability in distributive games (where agents compete over a fixed resource) yet are associated with richer exploration in integrative (cooperative) settings. Further, they suggest that findings obtained under English-only conditions may not generalize to other linguistic settings. These findings caution against English-only evaluation of LLMs and suggest that linguistically-grounded evaluation is essential for fair deployment.
\end{abstract}

\section{Introduction}

Negotiation is a fundamental form of social and economic interaction, requiring agents to reason strategically, balance self-interest with cooperation, and adapt behavior based on contextual and social cues \cite{lewis2017dealornodeal, he2018decoupling}. With the emergence of Large Language Models (LLMs), prior studies demonstrate that LLMs can engage in multi-turn bargaining and achieve non-trivial outcomes in competitive settings \cite{kwon2024effective, vaccaro2025ai}. Frameworks such as NegotiationArena \cite{bianchi2024negotiationarena} demonstrate that LLMs exhibit human-like negotiation behaviors, including anchoring and concession patterns. Nearly all studies conduct their evaluation exclusively in English, implicitly treating language as a neutral communication channel.

\noindent Extensive evidence from linguistics suggests that linguistic framing influences trust, cooperation, and strategic decision-making in human interactions \cite{hall1976beyond, brett2007negotiating}. If LLMs internalize language-conditioned patterns from training data, then the interaction language may systematically shape strategic behavior even when incentives remain fixed. This issue is particularly salient in multilingual contexts, where LLM performance is often framed as a matter of simple degradation \cite{dey2024bettertoask, singh2024indicgenbench}. However, recent inquiries suggest that language choice may fundamentally alter the structure of interaction \cite{tam-etal-2025-language, babel_effect_2024}. We investigate whether negotiation behavior is invariant to language in LLMs. \cite{price_of_thought_2025}.

\noindent To our knowledge, this study is the first to hold game rules, model parameters, and incentives constant while systematically varying linguistic framing across multi-agent negotiation games. We demonstrate that the interaction language acts as a non-trivial strategic prior, introducing variances in outcomes that persist across different model architectures. 
We demonstrate that a model's strategic intelligence is not a fixed attribute but is qualitatively altered by the language of interaction. Across 4,320 games spanning three canonical settings, we find statistically 
significant effects. In the \textbf{Ultimatum Game}, Gujarati and 
Punjabi framing significantly reduces acceptance rates relative to 
English, Punjabi produces significantly lower initial offers, and all 
Indic conditions yield significantly longer negotiations. In the \textbf{Buy-Sell Game}, Punjabi significantly 
extends negotiation length relative to English. In the \textbf{Resource Exchange Game}, all Indic 
languages significantly increase trade volume over English, while payoffs 
remain balanced.

\noindent These findings show that language is an active component of strategic reasoning rather than a passive medium. This work establishes that multilingual evaluation is a fundamental requirement for fair and robust AI deployment. We move beyond simple performance degradation narratives to show that language is associated with entirely different behavioral regimes. For developers and policymakers, these findings serve as a necessary caution: an LLM that is cooperative in English may become adversarial or sub-optimal in another language, even when provided with identical instructions. 

\section{Related Work}

\noindent Recent work has increasingly challenged the assumption that language is a neutral variable for large language models (LLMs). Beyond basic performance metrics, recent research has highlighted how the choice of language fundamentally alters a model's internal processing and alignment. \cite{tam-etal-2025-language} investigate the "English-as-a-hub" phenomenon, demonstrating that large reasoning models often default to English for internal chain-of-thought steps even when the input is in another language, which can lead to a significant performance-alignment trade-off. This linguistic contingency extends to the ethical domain; Agarwal et al. \cite{agarwal-etal-2024-ethical} show that moral value alignment and ethical reasoning in frontier models like GPT-4 are not universal but vary significantly depending on the prompt language. Similarly, the Babel Effect \cite{babel_effect_2024} demonstrates systematic performance disparities across languages, highlighting the English-centric nature of current LLM training and evaluation. These works establish that language influences reasoning accuracy and consistency, but primarily focus on static tasks.

\noindent In parallel, emerging work has begun to explore multilingual effects in interactive settings. \cite{price_of_thought_2025} examines how language impacts negotiation efficiency, reasoning cost, and outcome quality. This line of work suggests that multilingual reasoning introduces trade-offs between computational cost and performance. However, prior approaches largely treat language as a factor affecting efficiency or correctness, rather than as a variable that can fundamentally alter interaction.

\noindent A separate body of literature studies LLMs as negotiators. Subsequent work shows that cooperation, agreeableness, and persona conditioning significantly influence outcomes \cite{kwon2024effective, vaccaro2025ai}. Persona-based studies further highlight that lightweight contextual signals can strongly modulate negotiation strategies \cite{jeon2024mimicking,cohen2025bigfive}. However, these studies are predominantly conducted in English, implicitly treating language as invariant.

\noindent Finally, multilingual evaluation studies document persistent performance gaps across languages, particularly in low-resource settings \cite{dey2024bettertoask, singh2024indicgenbench}. While this literature frames multilingual differences primarily as bias or degradation, it does not examine whether language induces qualitatively different behaviors in interactive tasks.

\noindent Unlike prior work that focuses on reasoning accuracy or efficiency, we show that linguistic framing is a non-trivial factor influencing negotiation dynamics in LLMs, with effects that vary across task structure. Specifically, by holding models, incentives, and game structure constant, we present a controlled experimental framework to study the effect of interaction language in multi-agent LLM negotiation across three canonical games. This positions our work at the intersection of multilingual reasoning and multi-agent interaction, highlighting the need for evaluation frameworks that treat language as an active component of strategic reasoning rather than a passive medium.

\section{Hypotheses}
\label{sec:theory}

We generate two testable hypotheses and test them examining where LLM behavior aligns with or deviates from theory:

\noindent\textbf{H1 (Language-Mediated Strategy):} LLM negotiations are not language agnostic; instead, the choice of language systematically reshapes strategic behavior, social norm adherence, and equilibrium outcomes.

\noindent\textbf{H2 (Task Contingency):} Effects vary by game structure between distributive tasks (Ultimatum, Buy-Sell) vs integrative tasks (Resource Exchange).


\section{Methodology}
\label{sec:method}

We extend the \textbf{NegotiationArena} framework~\cite{bianchi2024negotiationarena}, which provides structured multi–agent negotiation games, turn–based dialogue control, and standardised evaluation protocols. By holding incentives, model parameters, and game structure constant, we ensure that observed behavioral differences are attributable to linguistic framing alone. All experiments were run across three core games included in the framework:

\noindent \textbf{BuySell Game:} P1 is the seller with a minimum acceptable price, and P2 is the buyer with a maximum willingness to pay.

\noindent \textbf{Ultimatum Game:} An asymmetric power negotiation game. P1 proposes a division of a fixed resource pool (e.g., 100 units). P2 may accept (both receive the proposed split) or reject (both receive zero). 

\noindent \textbf{Resource Exchange Game:} Each agent has access to a set of resources and a goal. For example, an agent has access to resources 25 Xs and 5 Ys. The agent might have the goal of maximizing its total resources. P1 initiates the first offer.


\subsection{System Prompts and Persona Design}
\label{subsec:prompts}

We design system prompts that assign each agent a specific linguistic identity. Our three primary linguistic framings are:
Hindi, Gujarati, Punjabi. All prompts explicitly forbid internal chain-of-thought, requiring only short rationale summaries, following the experimental setup of ~\cite{bianchi2024negotiationarena}. This constraint is necessary to isolate the model's immediate behavioral outputs and prevent the introduction of an uncontrolled variable: reasoning traces which could mask the direct influence of linguistic framing on decision-making. The persona prompts are: "You speak and bargain only in [language]. Negotiate accordingly." The choice to write system prompts in English was made to maintain consistency with the NegotiationArena framework ~\cite{bianchi2024negotiationarena}, which uses English-language instructions. We also run the games without any lingual prompting, providing an English baseline. All experiments were run

\subsection{Model Settings}
\label{subsec:models}

We evaluate a set of four multilingual LLMs, GPT-4o, GPT-3.5 Turbo, Claude-3-Haiku, Claude-3.5-Haiku. Temperature=0.7 and sampling settings are held constant across all linguistic conditions. Each game is repeated thirty times per condition to observe stable behavioural trends.

\subsection{Experimental Factors}
\label{subsec:conditions}

Experiments were conducted for the three games in a Model A vs Model B format ($A \neq B$) for four languages. Model A and Model B simply refer to assigning a Model A to Player 1(P1) and Model B to Player 2(P2). All ordered pairs of models were chosen. Each run logs: full dialogue, parsed offers or resource splits, final utilities, agreement/acceptance decisions. All combinations of experiments were run across thirty runs, with standardized logging of dialogues and offers.
Total experiments run $= 4(models) \times 3(othermodels) \times 4(languages) \times 30(runs) \times 3(games) = 4320$.

\subsection{Evaluation Principles and Metrics}
Following prior work on computational negotiation and multi-agent bargaining evaluation that emphasize that evaluating bargaining systems requires measuring not only whether agreements occur, but also how value is allocated and how negotiation unfolds over time, our evaluation metrics are grouped into three categories: \textbf{(i) Outcome Stability Metrics}, which measure whether negotiations successfully reach agreements \textit{(Acceptance Rate)}.\textbf{ (ii) Value Distribution Metrics}, which capture how negotiated resources are allocated between agents \textit{(Player Payoffs, Win Rate, Buyer/Seller Advantage)}. \textbf{(iii) Interaction Dynamics Metrics}, which characterize the negotiation process and strategic exploration \textit{(Conversation Rounds, Trade Volume).}

\noindent We adopt the four following objective negotiation metrics across all the three games:
\textbf{Acceptance Rate} measures the proportion of proposals accepted by P2. \textbf{Player Payoffs} capture final resource allocation for each player, summing all resources including exchanged items. \textbf{Win Rate (P1)} is the ratio of P1 wins to non-draw games, where a win is defined as having greater resources than the other player. \textbf{Conversation Rounds} counts negotiation turns before a final decision. Additionally, we adopt certain additional metrics specific to each game: 

\noindent\textbf{Ultimatum Game:} \textbf{Initial Offer} represents the average amount P1 offers to P2. 

\noindent \textbf{Buy-Sell Game:} \textbf{Buyer Advantage} is defined as the difference of the maximum amount the buyer is willing to pay and the actual trade price. \textbf{Seller Advantage} is defined as the difference between the actual trade price and the minimum amount the seller is willing to sell at. 

\noindent \textbf{Resource Exchange Game:} \textbf{Trade Volume} measures the number of resources that have exchanged hands.

\noindent For each behavior, metrics were aggregated across all ordered model combinations using raw game data: rates were calculated from total counts, while payoffs, offers, and rounds were computed as means and standard deviations from concatenated arrays of individual outcomes.

\subsection{Statistical Testing}
\label{subsec:stattesting}

To assess whether language behavior significantly affects negotiation outcomes, we apply non-parametric statistical tests across all language conditions. For continuous metrics (trade volume, payoffs, and negotiation rounds), we use the Kruskal-Wallis H-test for overall differences, followed by pairwise Mann-Whitney U tests. For binary outcomes (acceptance rate and win rate), we use chi-square tests with pairwise two-proportion $z$-tests. All pairwise $p$-values are corrected using the Benjamini-Hochberg false discovery rate procedure. We report significance at three levels: $^{*}$($p_{\text{corr}} < 0.05$), $^{**}$($p_{\text{corr}} < 0.01$), $^{***}$($p_{\text{corr}} < 0.001$), with non-significant results denoted \textit{ns}.

\subsection{Language Compliance Analysis}
\label{subsec:langcomply}

To verify linguistic compliance, we applied the AI4Bharat IndicNLP language identification model \cite{madhani-etal-2023-bhasa}, which uses FastText embeddings pretrained on a large-scale monolingual Indic corpus covering 22 Indian languages, to all conversation turns across all experimental conditions. We report adherence rates 
and average confidence scores in Appendix \ref{sec:langcompliance}. Overall, adherence rates are high across all verified languages and games (89.65\%--97.85\%), with near-perfect 
confidence scores ($\geq$0.966), indicating that models reliably 
produced output in the instructed language.

\section{Results and Analysis}

We compare the Baseline English condition with multiple Indian language contexts (Gujarati, Hindi, Punjabi) in all three games. All results are supported by non-parametric statistical testing (Kruskal-Wallis, Mann-Whitney U, and chi-square tests with Benjamini-Hochberg correction); full test statistics and p-values are reported in Appendix \ref{sec:statanaly}.

\subsection{Ultimatum Game Results}
\begin{figure*}[h]
    \centering
    \includegraphics[width=1\linewidth]{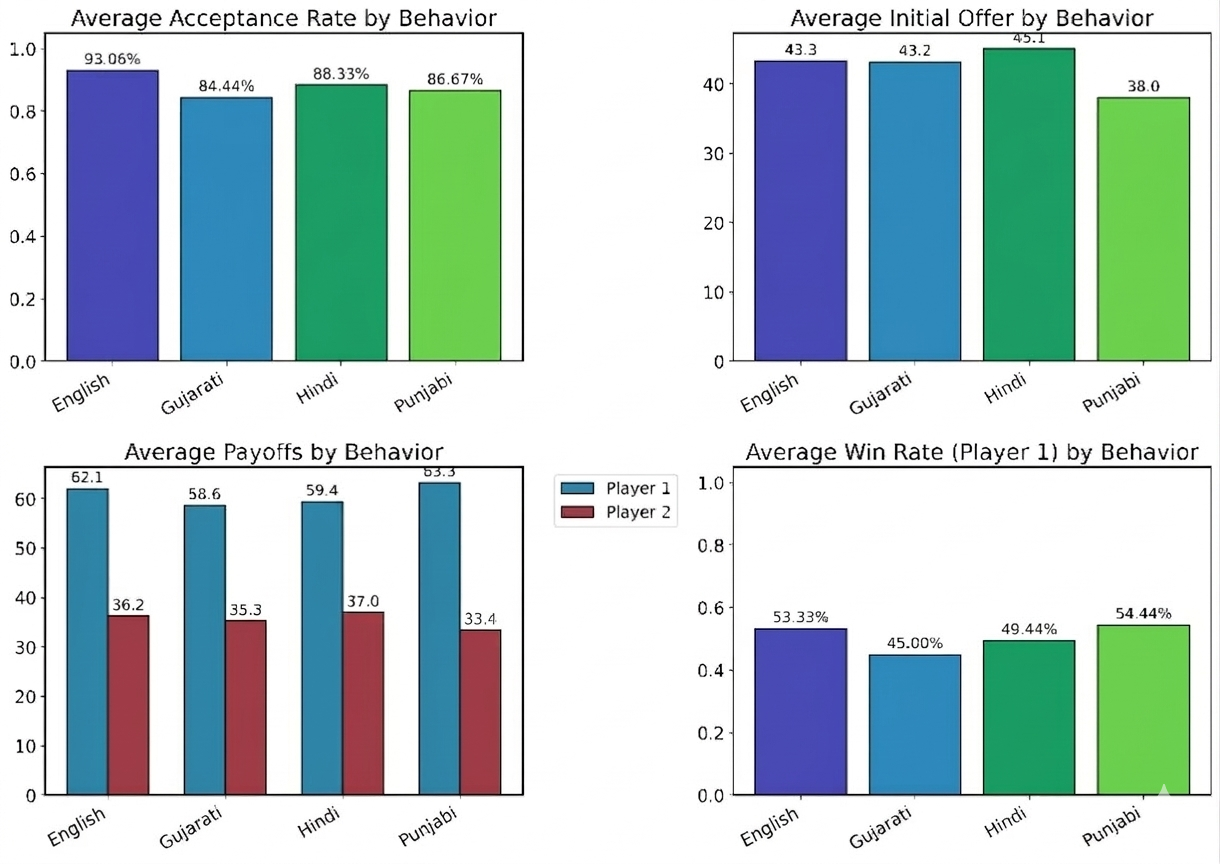}
    \caption{Ultimatum Game Language Comparison showing average (a) acceptance rates, (b) average initial offer, (c) payoffs, (d) win rates (P1).}
    \label{fig:ulti_lang}
\end{figure*}

Results are shown in Figure \ref{fig:ulti_lang} and Table \ref{tab:ultimatum_summary}.

\begin{table*}[htbp]
\centering
\resizebox{\textwidth}{!}{%
\small
\begin{tabular}{lcccccc}
\toprule
Language & Acceptance Rate & Initial Offer & P1 Payoff & P2 Payoff & P1 Win Rate & Conversation Rounds \\
\midrule
English & \textbf{93.06\% $\pm$ 25.42\%} & 43.31 $\pm$ 11.26 & 62.09 $\pm$ 19.57 & 36.25 $\pm$ 18.44 & 53.33\% $\pm$ 49.89\% & \textbf{2.55 $\pm$ 1.03} \\
Gujarati & 84.44\% $\pm$ 36.24\% & 43.20 $\pm$ 13.59 & 58.64 $\pm$ 26.48 & 35.25 $\pm$ 23.63 & 45.00\% $\pm$ 49.75\% & 3.10 $\pm$ 1.41 \\
Hindi    & 88.33\% $\pm$ 32.10\% & \textbf{45.08 $\pm$ 13.51} & 59.42 $\pm$ 22.36 & \textbf{36.97 $\pm$ 20.47} & 49.44\% $\pm$ 50.00\% & 3.04 $\pm$ 1.37 \\
Punjabi  & 86.67\% $\pm$ 33.99\% & 37.96 $\pm$ 15.36 & \textbf{63.29 $\pm$ 24.45} & 33.38 $\pm$ 22.32 & \textbf{54.44\% $\pm$ 49.80\%} & 3.14 $\pm$ 1.48 \\
\bottomrule
\end{tabular}
}
\caption{Metrics for \textbf{Ultimatum Game} aggregated across all model combinations (mean $\pm$ std).}
\label{tab:ultimatum_summary}
\end{table*}

\subsubsection{Baseline Language (English)}
The English condition exhibits stable and efficient negotiation dynamics. It achieves a high acceptance rate ($93.1\%$), with an average initial offer of 43.31 ZUP. P1 earns 62.09 ZUP on average, while P2 receives 36.25 ZUP, indicating a moderately P1-favored outcome. Conversations remain short (2.55 rounds on average).

\subsubsection{Multilingual Outcome Variation}

\noindent (1) \textbf{Acceptance and Cooperation.} Acceptance rates decrease relative to the English baseline in all other languages: Gujarati ($84.4\%$), Hindi ($88.3\%$), and Punjabi ($86.7\%$). The reductions for Gujarati ($p_{\text{corr}} = 0.0015^{**}$) and Punjabi ($p_{\text{corr}} = 0.0135^{*}$) are statistically significant, while the Hindi difference does not remain significant after correction.

\noindent (2) \textbf{Payoff Balance and Efficiency.} P1 payoff remains broadly comparable across languages, with no statistically significant differences relative to the baseline after correction. A notable exception is a significant difference between Gujarati and Punjabi ($p_{\text{corr}} = 0.0406^{*}$), where Punjabi yields higher P1 payoff. P2 payoff does not exhibit statistically significant variation across conditions.

\noindent (3) \textbf{Deviation from English Baseline.} The most pronounced deviations from the baseline arise in initial offer behavior and interaction length. Punjabi produces substantially lower initial offers (mean $37.96$), significantly below English, Gujarati, and Hindi. Additionally, all non-English conditions result in longer negotiations, with significantly more turns than the baseline, indicating increased interaction before agreement.

\subsubsection{Evaluating Hypotheses}
The results provide clear evidence in support of H1 (Language-Mediated Strategy). Several outcome dimensions vary systematically with language. Acceptance rates are significantly lower in Gujarati and Punjabi relative to the English baseline ($p_{\text{corr}} = 0.0015^{**}$ and $0.0135^{*}$, respectively), while Punjabi exhibits substantially lower initial offers compared to all other conditions (all pairwise $p_{\text{corr}} < 0.001^{***}$). In addition, all non-English conditions show significantly longer negotiations (all $p_{\text{corr}} < 0.001^{***}$).

\subsubsection{Evaluating Prompt Sensitivity}

To verify that our findings are not an artifact of the specific persona prompt wording, we replicate the Ultimatum Game under three variants of the prompts: two semantically equivalent phrasings of the original prompt (A1--A2) and a native-script prompt where the persona directive is written directly in the target language (A3), keeping all other conditions constant. While absolute values vary only modestly across variants, the relative behavior replicates consistently: acceptance rates remained high and stable under English (91.9--93.3\%), reduced acceptance rates in Gujarati and Punjabi, Punjabi produces the lowest initial offers across all runs (37.4--41.1 vs.\ 41.4--46.4 for other languages), and the P1-favored payoff asymmetry persists throughout. Notably, A3, the native-script condition, does not produce systematically different trends from the English-phrased variants, suggesting that the observed effects are driven by the interaction language itself rather than the script or the phrasing of the prompt. The one notable source of variance across prompts is win rate under Punjabi, where A1 yields an elevated P1 win rate of 70.8\% compared to 52.2--56.7\% in other variants, suggesting this metric carries more noise. Comprehensive per prompt variant results are reported in Appendix~\ref{sec:promptablation}.

\begin{figure*}[h]
    \centering
    \includegraphics[width=1\linewidth]{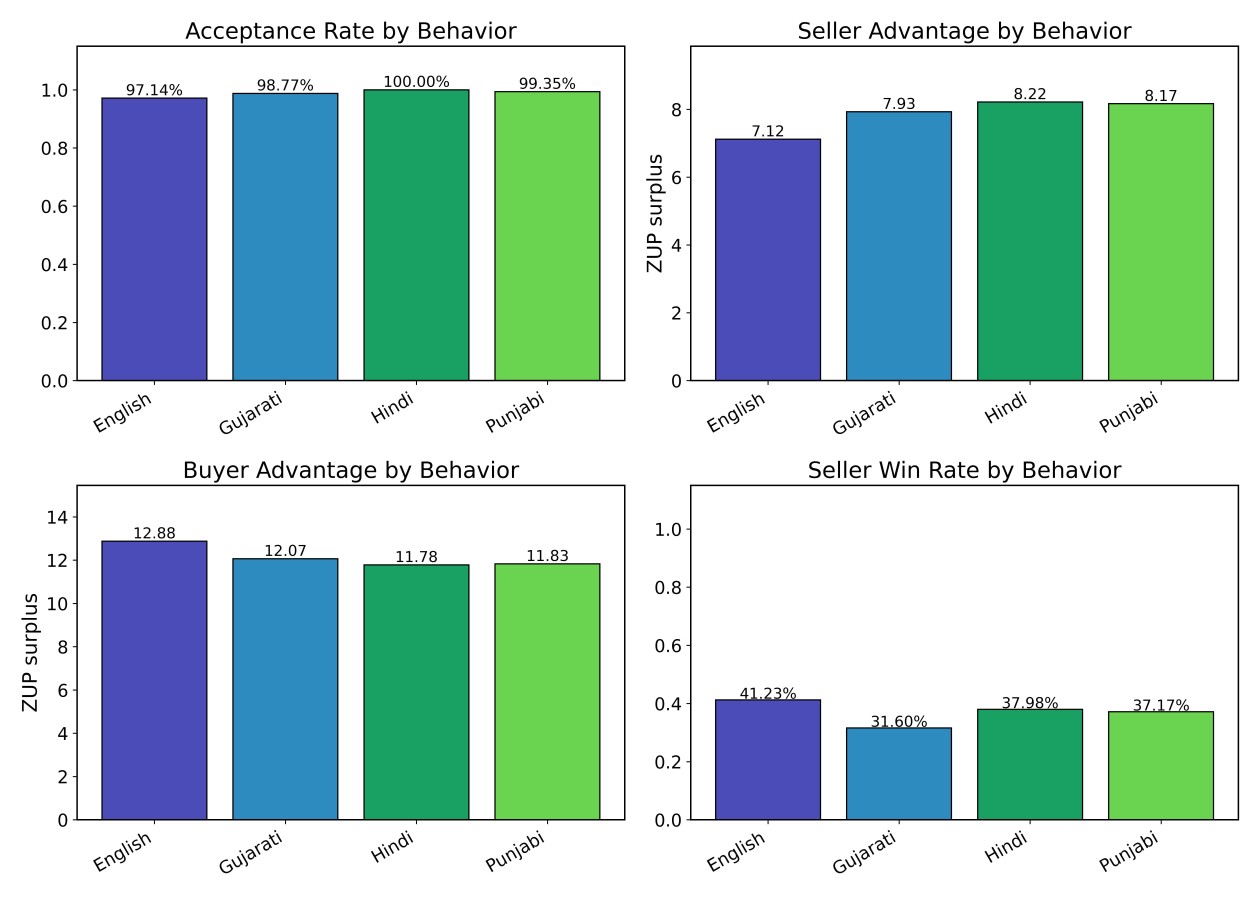}
    \caption{Buy Sell Game Language Comparison showing average (a) acceptance rates, (b) seller advantages, (c)  buyer advantages, and (d) win rates across different linguistic behaviors.}
    \label{fig:buysell_lang}
\end{figure*}
\begin{table*}[htbp]
\resizebox{\textwidth}{!}{%
\centering
\small
\begin{tabular}{lccccc}
\toprule
Language & Acceptance Rate & Seller Advantage & Buyer Advantage & Conversation Rounds & P1 Win Rate \\
\midrule
English  & 97.14\% $\pm$ 16.68\% & 7.12 $\pm$ 12.39 & \textbf{12.88 $\pm$ 12.39} & 3.09 $\pm$ 1.90 & \textbf{41.23\%} \\
Gujarati & 98.77\% $\pm$ 11.04\% & 7.93 $\pm$ 9.82  & 12.07 $\pm$ 9.82  & \textbf{3.31 $\pm$ 1.92} & 31.60\% \\
Hindi    & \textbf{100.00\% $\pm$ 0.00\%} & \textbf{8.22 $\pm$ 11.57} & 11.78 $\pm$ 11.57 & 3.04 $\pm$ 1.46 & 37.98\% \\
Punjabi  & 99.35\% $\pm$ 8.05\% & 8.17 $\pm$ 10.86 & 11.83 $\pm$ 10.86 & 3.42 $\pm$ 1.87 & 37.17\% \\
\bottomrule
\end{tabular}
}
\caption{Metrics for \textbf{Buy-Sell Game} aggregated across all model combinations (mean $\pm$ std).}
\label{tab:buy_sell_language_summary}
\end{table*}




\begin{figure*}[h]
    \centering
    \includegraphics[width=1\linewidth]{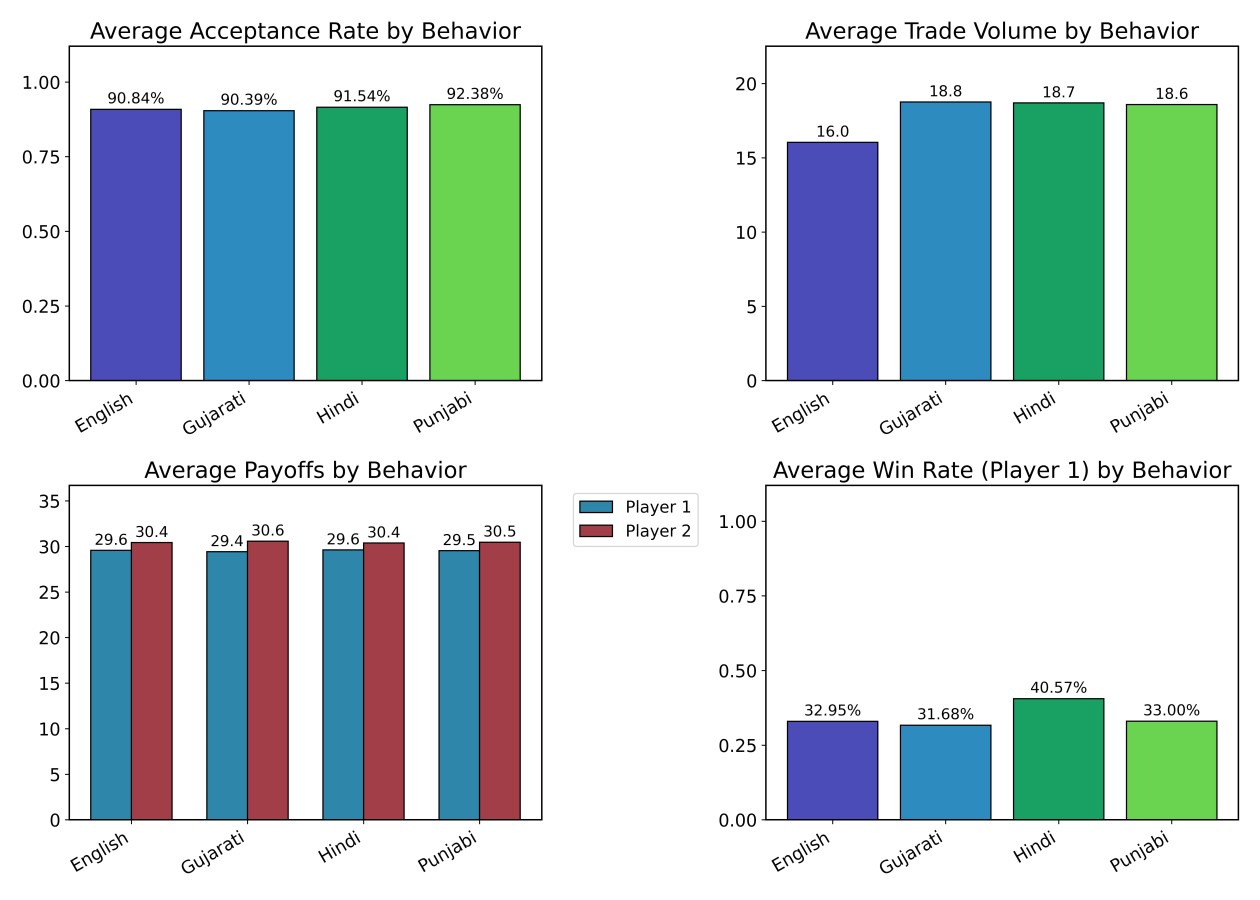}
    \caption{Resource Exchange Game Language Comparison showing average (a) acceptance rates, (b) trade volume, (c) payoffs, and (d) win rates (P1) across different linguistic behaviors.}
    \label{fig:trading_lang}
\end{figure*}
\begin{table*}[htbp]
\centering
\resizebox{\textwidth}{!}{%
\small
\begin{tabular}{lcccccc}
\toprule
Language & Acceptance Rate (\%) & Trade Volume & P1 Payoff & P2 Payoff & P1 Win Rate (\%) & Conversation Rounds \\
\midrule
English  & 90.84 $\pm$ 28.90 & 16.05 $\pm$ 6.77 & 29.57 $\pm$ 2.57 & 30.43 $\pm$ 2.57 & 32.95 & 3.10 $\pm$ 1.42 \\
Gujarati & 90.39 $\pm$ 29.52 & \textbf{18.77 $\pm$ 7.99} & 29.42 $\pm$ 2.76 & \textbf{30.58 $\pm$ 2.76} & 31.68 & \textbf{3.28 $\pm$ 1.41} \\
Hindi    & 91.54 $\pm$ 27.88 & 18.70 $\pm$ 6.87 & \textbf{29.63 $\pm$ 2.79} & 30.37 $\pm$ 2.79 & \textbf{40.57} & 3.27 $\pm$ 1.41 \\
Punjabi  & \textbf{92.38 $\pm$ 26.57} & 18.59 $\pm$ 6.70 & 29.53 $\pm$ 2.64 & 30.47 $\pm$ 2.64 & 33.00 & 3.15 $\pm$ 1.40 \\
\bottomrule
\end{tabular}
}
\caption{Metrics for \textbf{Resource Exchange} aggregated across all model combinations (mean $\pm$ std).}
\label{tab:trading_behavior_summary}
\end{table*}

\subsection{Buy-Sell Game Results}
Results are shown in Figure \ref{fig:buysell_lang} and Table \ref{tab:buy_sell_language_summary}.

\subsubsection{Baseline Language (English)}

English yields a high acceptance rate ($97.14\%$) with seller advantage (mean $7.12$, std $12.39$) lower than buyer advantage (mean $12.88$, std $12.39$). The seller win rate remains modest at $41.23\%$, and negotiations conclude in an average of $3.09$ rounds (std $1.90$).

\subsubsection{Multilingual Outcome Variation}

All non-English languages achieve near-perfect acceptance rates, with Hindi reaching $100\%$ agreement and Gujarati ($98.77\%$) and Punjabi ($99.35\%$) closely following.

\noindent While average seller advantage increases slightly in Gujarati ($7.93$), Hindi ($8.22$), and Punjabi ($8.17$) relative to English, and buyer advantage correspondingly decreases, these differences remain small in magnitude.

\noindent More pronounced variation emerges in negotiation dynamics. Punjabi produces the highest average number of negotiation rounds ($3.42$, std $1.87$), followed by Gujarati ($3.31$) and English ($3.09$), with Hindi lowest ($3.04$). The increase in negotiation rounds for Punjabi relative to English is statistically significant ($p\_{\text{corr}} = 0.0012^{**}$), and Punjabi also exceeds Hindi ($p\_{\text{corr}} = 0.0468^*$).

\subsubsection{Evaluating Hypotheses}

The primary statistically supported effect concerns negotiation length. Punjabi is consistently associated with longer negotiations compared to English ($p_{\text{corr}} = 0.0012^{**}$), providing clear evidence that language choice can influence interaction dynamics.

\noindent In contrast, differences in seller and buyer advantage across languages are modest and do not yield statistically significant pairwise contrasts after correction. Accordingly, the data do not provide direct evidence for systematic shifts in surplus allocation corresponding to H1 or H2.





\subsection{Resource Exchange Game Results}
Results are shown in Figure \ref{fig:trading_lang} and Table \ref{tab:trading_behavior_summary}.

\subsubsection{Baseline Language (English)}

English yields a $90.84\%$ acceptance rate, indicating that LLM agents consistently reach agreement under this condition. However, English produces the lowest average trade volume ($16.05$).

\noindent Payoff distributions in English are closely balanced: P1 achieves an average payoff of $29.53$ and P2 $30.47$, with no statistically significant differences relative to other languages after correction. Similarly, P1 wins $33.0\%$ of non-tied games, a rate that does not differ significantly across languages.

\subsubsection{Multilingual Outcome Variation}

Trade volume exhibits clear and statistically significant variation. Gujarati ($18.77$), Hindi ($18.70$), and Punjabi ($18.59$) all yield substantially higher average trade volumes than English ($16.05$). Pairwise comparisons confirm that all non-English languages differ significantly from English ($p_{\text{corr}}=0.0001^{***}$), while differences among Gujarati, Hindi, and Punjabi are not statistically significant.

\noindent Other metrics remain largely stable across languages. P1 and P2 payoffs show no significant pairwise differences after correction, with means tightly clustered around $30$. Negotiation rounds exhibit minor variation (English: $3.10$; Gujarati: $3.28$; Hindi: $3.27$; Punjabi: $3.15$), but none of the comparisons reach statistical significance. Similarly, P1 win rates vary modestly (ranging from $31.7\%$ to $40.6\%$).

\subsubsection{Evaluating Hypotheses}

The results provide partial support for \textbf{H1 (Language-Mediated Strategy)}. Language choice systematically affects trade volume, with all non-English conditions inducing significantly higher exchange levels than English. However, other key outcomes - including payoffs, negotiation length, and win rates remain invariant across languages after correction, indicating that the effect of language may be selective rather than global.





\subsection{Model-Specific Performance}
\label{sec:model-specific}

As the primary focus of this work is on language effects in negotiation,
model-specific results are discussed briefly here; detailed heatmaps are
provided in Appendix~\ref{sec:visualizations}.

\paragraph{Ultimatum Game.}
GPT-4o exhibits asymmetric performance across roles: as P2 it is
comparatively difficult to win against, whereas as P1 it achieves a lower
win rate than GPT-3.5 under comparable conditions. Notably, GPT-4o
consistently provides its negotiation rationale in the interaction language,
while the other models evaluated often default to English regardless of the
instructed language. This difference in response behaviour coincides with
the observed performance variation across conditions.

\paragraph{Buy-Sell Game.}
The Buy-Sell Game exposes large role-dependent asymmetries that interact
with language choice. In the English baseline, GPT-4o as seller (P1) achieves a seller advantage
of 19.1--20.0, while GPT-3.5 in the same role records only 0.2--0.9---a
gap of nearly 20 points. Conversely, as buyer (P2), GPT-3.5 secures
extreme advantages of 26.5--28.1, indicating systematic over-concession
when selling and over-extraction when buying. GPT-4o as buyer (P1) yields
negligible buyer advantages and, conversely, consistently provides the
largest seller advantages as P1 across all languages.

\noindent Under Indic language conditions, these asymmetries are attenuated but not
eliminated. In Gujarati, GPT-3.5's seller advantage as P1 improves to
$-$2.0 to $-$0.4, while its buyer advantage falls to 20.4--22.0. GPT-4o
maintains a seller advantage of 13.3--16.5 in Gujarati, preserving its
dominant position. This pattern holds across all tested languages: GPT-4o
consistently achieves the largest seller advantages as P1, and GPT-3.5
consistently achieves the largest buyer advantages as P1, across all
languages. Taken together, these results suggest that linguistic framing
may attenuate but does not eliminate capacity-driven asymmetries between
models.

\section{Conclusion}

This study explores how linguistic framing influences the dynamics of negotiation in Large Language Models across three game-theoretic environments. Our observations suggest that language choice may not be a neutral factor in strategic interactions; rather, it correlates with shifts in negotiation outcomes and surplus allocation, particularly within the Indic language contexts studied. While we find that certain linguistic framings are associated with different patterns of stability or exploration compared to English, these effects appear to be highly task-contingent. The contrast between the Ultimatum and Resource Exchange findings provides support for H2: language effects are task-contingent, reducing stability in distributive games while increasing exploration in integrative ones. These results highlight the potential limitations of evaluating strategic reasoning solely through an English-centric lens. Our findings suggest that incorporating a broader range of languages into the evaluation of LLM agents is an important step toward a more comprehensive understanding of their capabilities in diverse linguistic settings as LLMs deploy globally in commercial and interpersonal contexts, with direct implications for fairness and equitable deployment. Any benchmark or evaluation framework for LLM negotiation or strategic reasoning that is English-only should be considered incomplete, and multilingual evaluation should be treated as a methodological requirement rather than an optional extension.

\section{Limitations}

Our findings should be interpreted carefully. The behaviors exhibited by LLM agents do not constitute evidence about real human negotiation practices or cultural norms - differences reflect patterns learned from training corpora, not properties of languages or their speakers. Our language framings use culturally associated labels without incorporating human participants or sociolinguistic context; any apparent alignment with stereotypes should be understood as an artifact of representation learning, analogous to well-documented biases in word embeddings. We deliberately avoid normative claims and do not endorse any interpretation that attributes these behaviors to real-world groups. Our system prompts are written in English to maintain consistency with NegotiationArena, which may partially attenuate the linguistic signal being measured. Third, our analysis covers limited games and languages. While spanning distributive and integrative settings, these games lack the richness of real-world negotiation (long-term relationships, incomplete information).
We also examine only model-model interaction across three games, abstracting away from human-AI dynamics, richer incomplete-information settings, and realistic code-switching such as Hinglish. In addition, a more specific lingual system prompt could be curated; in our results we observed some models would provide their reason for the trade in the respective language while others would stick to English, it would be valuable to evaluate this variable as well. While our results provide a proof-of-concept that linguistic framing functions as a significant behavioral prior, whether these effects generalize to typologically distant languages such as Mandarin or Arabic remains an open question for future work.

\noindent Despite these limitations, our results provide valuable evidence that language-conditioned representations influence strategic interaction in LLMs, underscoring the need for multilingual evaluation in socially sensitive domains.

\section{Acknowledgments}

We thank the creators of Negotiation Arena \cite{bianchi2024negotiationarena} for making their setup publicly available. We are also grateful for access to API platforms that made this evaluation possible. The authors also wish to acknowledge the usage of ChatGPT and Claude in improving the presentation and grammar of the paper. The paper remains an accurate representation of the authors' underlying contributions.

\bibliography{custom}

@inproceedings{lewis2017dealornodeal,
    title = "Deal or No Deal? End-to-End Learning of Negotiation Dialogues",
    author = "Lewis, Mike  and
      Yarats, Denis  and
      Dauphin, Yann  and
      Parikh, Devi  and
      Batra, Dhruv",
    editor = "Palmer, Martha  and
      Hwa, Rebecca  and
      Riedel, Sebastian",
    booktitle = "Proceedings of the 2017 Conference on Empirical Methods in Natural Language Processing",
    month = sep,
    year = "2017",
    address = "Copenhagen, Denmark",
    publisher = "Association for Computational Linguistics",
    url = "https://aclanthology.org/D17-1259/",
    doi = "10.18653/v1/D17-1259",
    pages = "2443--2453"
}

@inproceedings{he2018decoupling,
    title = "Decoupling Strategy and Generation in Negotiation Dialogues",
    author = "He, He  and
      Chen, Derek  and
      Balakrishnan, Anusha  and
      Liang, Percy",
    editor = "Riloff, Ellen  and
      Chiang, David  and
      Hockenmaier, Julia  and
      Tsujii, Jun{'}ichi",
    booktitle = "Proceedings of the 2018 Conference on Empirical Methods in Natural Language Processing",
    month = oct # "-" # nov,
    year = "2018",
    address = "Brussels, Belgium",
    publisher = "Association for Computational Linguistics",
    url = "https://aclanthology.org/D18-1256/",
    doi = "10.18653/v1/D18-1256",
    pages = "2333--2343",
}

@inproceedings{bianchi2024negotiationarena,
    author = {Bianchi, Federico and Chia, Patrick John and Yuksekgonul, Mert and Tagliabue, Jacopo and Jurafsky, Dan and Zou, James},
    title = {How well can {LLM}s negotiate? {NegotiationArena} platform and analysis},
    year = {2024},
    url = {https://arxiv.org/abs/2402.05863},
    publisher = {JMLR.org},
    booktitle = {Proceedings of the 41st International Conference on Machine Learning},
    articleno = {158},
    numpages = {17},
    location = {Vienna, Austria},
    series = {ICML'24}
}

@inproceedings{kwon2024effective,
    title = "Are {LLM}s Effective Negotiators? Systematic Evaluation of the Multifaceted Capabilities of {LLM}s in Negotiation Dialogues",
    author = "Kwon, Deuksin  and
      Weiss, Emily  and
      Kulshrestha, Tara  and
      Chawla, Kushal  and
      Lucas, Gale  and
      Gratch, Jonathan",
    editor = "Al-Onaizan, Yaser  and
      Bansal, Mohit  and
      Chen, Yun-Nung",
    booktitle = "Findings of the Association for Computational Linguistics: EMNLP 2024",
    month = nov,
    year = "2024",
    address = "Miami, Florida, USA",
    publisher = "Association for Computational Linguistics",
    url = "https://aclanthology.org/2024.findings-emnlp.310/",
    doi = "10.18653/v1/2024.findings-emnlp.310",
    pages = "5391--5413"
}

@misc{vaccaro2025ai,
      title = "Advancing {AI} Negotiations: New Theory and Evidence from a Large-Scale Autonomous Negotiations Competition",
      author = "Vaccaro, Michelle  and
        Caosun, Michael  and
        Ju, Harang  and
        Aral, Sinan  and
        Curhan, Jared R.",
      year = "2025",
      eprint = "2503.06416",
      archivePrefix = "arXiv",
      primaryClass = "cs.AI",
      url = "https://arxiv.org/abs/2503.06416",
}

@misc{cohen2025bigfive,
      title = "Exploring Big Five Personality and {AI} Capability Effects in {LLM}-Simulated Negotiation Dialogues",
      author = "Cohen, Myke C.  and
        Su, Zhe  and
        Kao, Hsien-Te  and
        Nguyen, Daniel  and
        Lynch, Spencer  and
        Sap, Maarten  and
        Volkova, Svitlana",
      year = "2025",
      eprint = "2506.15928",
      archivePrefix = "arXiv",
      primaryClass = "cs.AI",
      url = "https://arxiv.org/abs/2506.15928",
}

@inproceedings{jeon2024mimicking,
    title = "Mimicking Human Emotions: Persona-Driven Behavior of {LLM}s in the {`Buy and Sell'} Negotiation Game",
    author = "Jeon, Mingyu  and
      Suh, Jae Young",
    booktitle = "Language Gamification - NeurIPS 2024 Workshop",
    year = "2024",
    url = "https://openreview.net/forum?id=j7RkeNqSDo"
}

@misc{dey2024bettertoask,
      title = "Better to Ask in {English}: Evaluation of Large Language Models on {English}, Low-resource and Cross-Lingual Settings",
      author = "Dey, Krishno  and
        Tarannum, Prerona  and
        Hasan, Md. Arid  and
        Razzak, Imran  and
        Naseem, Usman",
      year = "2024",
      eprint = "2410.13153",
      archivePrefix = "arXiv",
      primaryClass = "cs.CL",
      url = "https://arxiv.org/abs/2410.13153",
}

@inproceedings{singh2024indicgenbench,
    title = "{I}ndic{G}en{B}ench: A Multilingual Benchmark to Evaluate Generation Capabilities of {LLM}s on {I}ndic Languages",
    author = "Singh, Harman  and
      Gupta, Nitish  and
      Bharadwaj, Shikhar  and
      Tewari, Dinesh  and
      Talukdar, Partha",
    editor = "Ku, Lun-Wei  and
      Martins, Andre  and
      Srikumar, Vivek",
    booktitle = "Proceedings of the 62nd Annual Meeting of the Association for Computational Linguistics (Volume 1: Long Papers)",
    month = aug,
    year = "2024",
    address = "Bangkok, Thailand",
    publisher = "Association for Computational Linguistics",
    url = "https://aclanthology.org/2024.acl-long.595/",
    doi = "10.18653/v1/2024.acl-long.595",
    pages = "11047--11073"
}

@book{brett2007negotiating,
    title = "Negotiating Globally: How to Negotiate Deals, Resolve Disputes, and Make Decisions Across Cultural Boundaries",
    author = "Brett, Jeanne M.",
    year = "2007",
    publisher = "John Wiley {\&} Sons"
}

@book{hall1976beyond,
    title = "Beyond Culture",
    author = "Hall, Edward T.",
    year = "1976",
    publisher = "Anchor"
}

@article{babel_effect_2024,
    title = "The {Babel} Effect: Analyzing Multilingual Performance Discrepancies in Large Language Models",
    author = "Jha, Basab",
    journal = "Engineering and Applied Sciences Journal",
    year = "2024"
}

@article{price_of_thought_2025,
    title = "The Price of Thought: A Multilingual Analysis of Reasoning, Performance, and Cost of Negotiation in Large Language Models",
    author = "Hakimov, Sherzod  and
      Bernard, Roland  and
      Leiber, Tim  and
      Osswald, Karl  and
      Richert, Kristina  and
      Yang, Ruilin  and
      Bernardi, Raffaella  and
      Schlangen, David",
    year = "2025"
}

@inproceedings{madhani-etal-2023-bhasa,
    title = "Bhasa-Abhijnaanam: Native-script and Romanized Language Identification for 22 {I}ndic Languages",
    author = "Madhani, Yash  and
      Khapra, Mitesh M.  and
      Kunchukuttan, Anoop",
    editor = "Rogers, Anna  and
      Boyd-Graber, Jordan  and
      Okazaki, Naoaki",
    booktitle = "Proceedings of the 61st Annual Meeting of the Association for Computational Linguistics (Volume 2: Short Papers)",
    month = jul,
    year = "2023",
    address = "Toronto, Canada",
    publisher = "Association for Computational Linguistics",
    url = "https://aclanthology.org/2023.acl-short.71/",
    doi = "10.18653/v1/2023.acl-short.71",
    pages = "816--826"
}

@inproceedings{agarwal-etal-2024-ethical,
    title = "Ethical Reasoning and Moral Value Alignment of {LLM}s Depend on the Language We Prompt Them in",
    author = "Agarwal, Utkarsh  and
      Tanmay, Kumar  and
      Khandelwal, Aditi  and
      Choudhury, Monojit",
    booktitle = "Proceedings of the 2024 Joint International Conference on Computational Linguistics, Language Resources and Evaluation (LREC-COLING 2024)",
    month = may,
    year = "2024",
    address = "Torino, Italia",
    publisher = "ELRA and ICCL",
    url = "https://aclanthology.org/2024.lrec-main.560/",
    pages = "6330--6340"
}

@article{tam-etal-2025-language,
    title = "Language Matters: How Do Multilingual Input and Reasoning Paths Affect Large Reasoning Models?",
    author = "Tam, Zhi Rui  and
      Wu, Cheng-Kuang  and
      Chiu, Yu Ying  and
      Lin, Chieh-Yen  and
      Chen, Yun-Nung  and
      Lee, Hung-yi",
    journal = "arXiv preprint arXiv:2505.17407",
    year = "2025",
    url = "https://arxiv.org/abs/2505.17407"
}

\clearpage
\appendix
\onecolumn
\section{Visualizations}
\label{sec:visualizations}


\begin{center}
  \includegraphics[width=\textwidth,height=0.9\textheight,keepaspectratio]{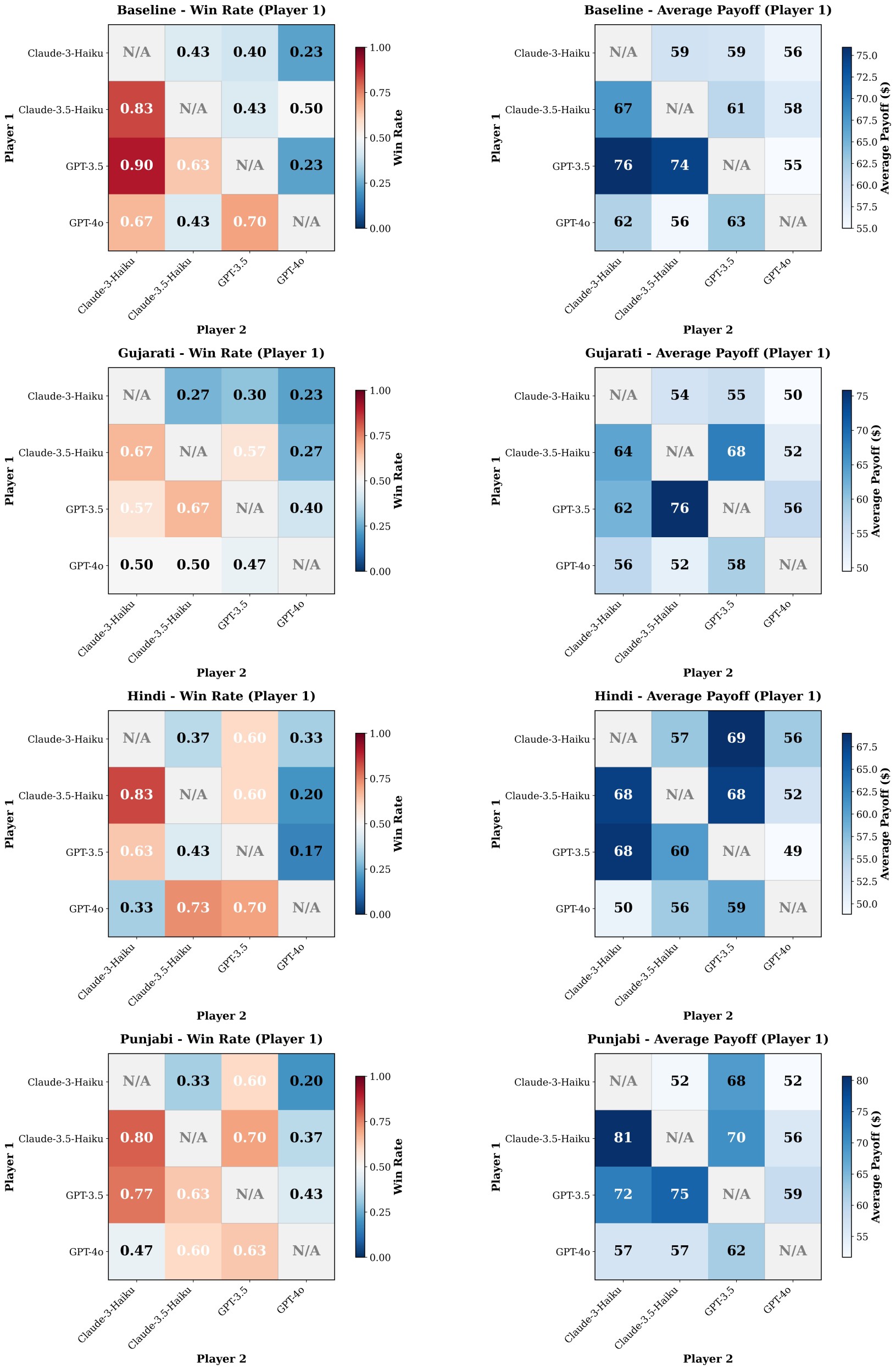}
\end{center}
\captionof{figure}{Heatmaps for \textbf{Ultimatum Game} comparing (a) win rate and (b) payoff for model combinations for all languages.}
\label{fig:ulti_heat}
\clearpage


\begin{center}
  \includegraphics[width=\textwidth,height=0.9\textheight,keepaspectratio]{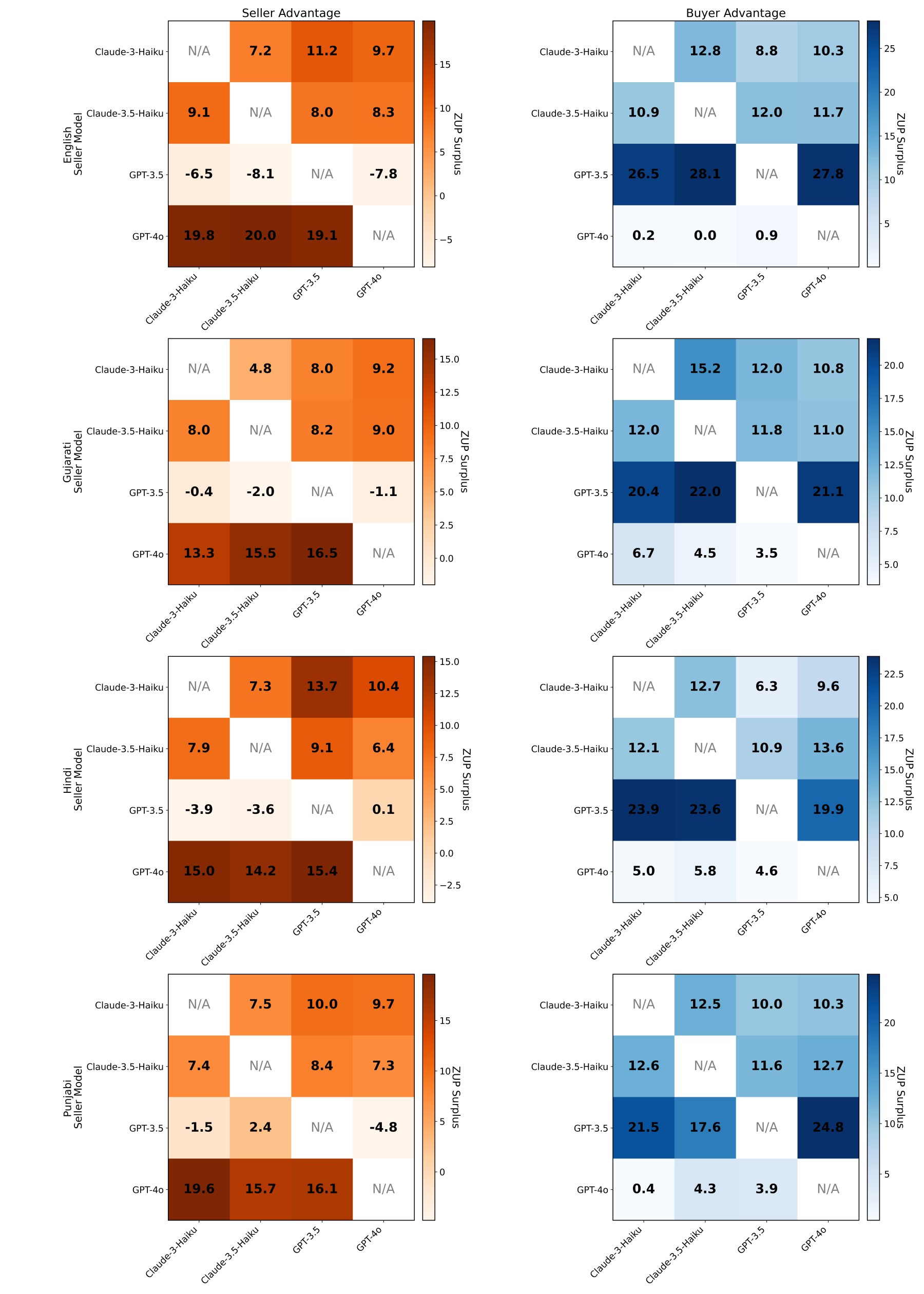}
\end{center}
\captionof{figure}{Heatmaps for \textbf{Buy-Sell Game} comparing (a) Seller advantage and (b) Buyer Advantage for model combinations for all languages.}
\label{fig:buysell_allheatmaps}
\clearpage


\begin{center}
  \includegraphics[width=\textwidth,height=0.9\textheight,keepaspectratio]{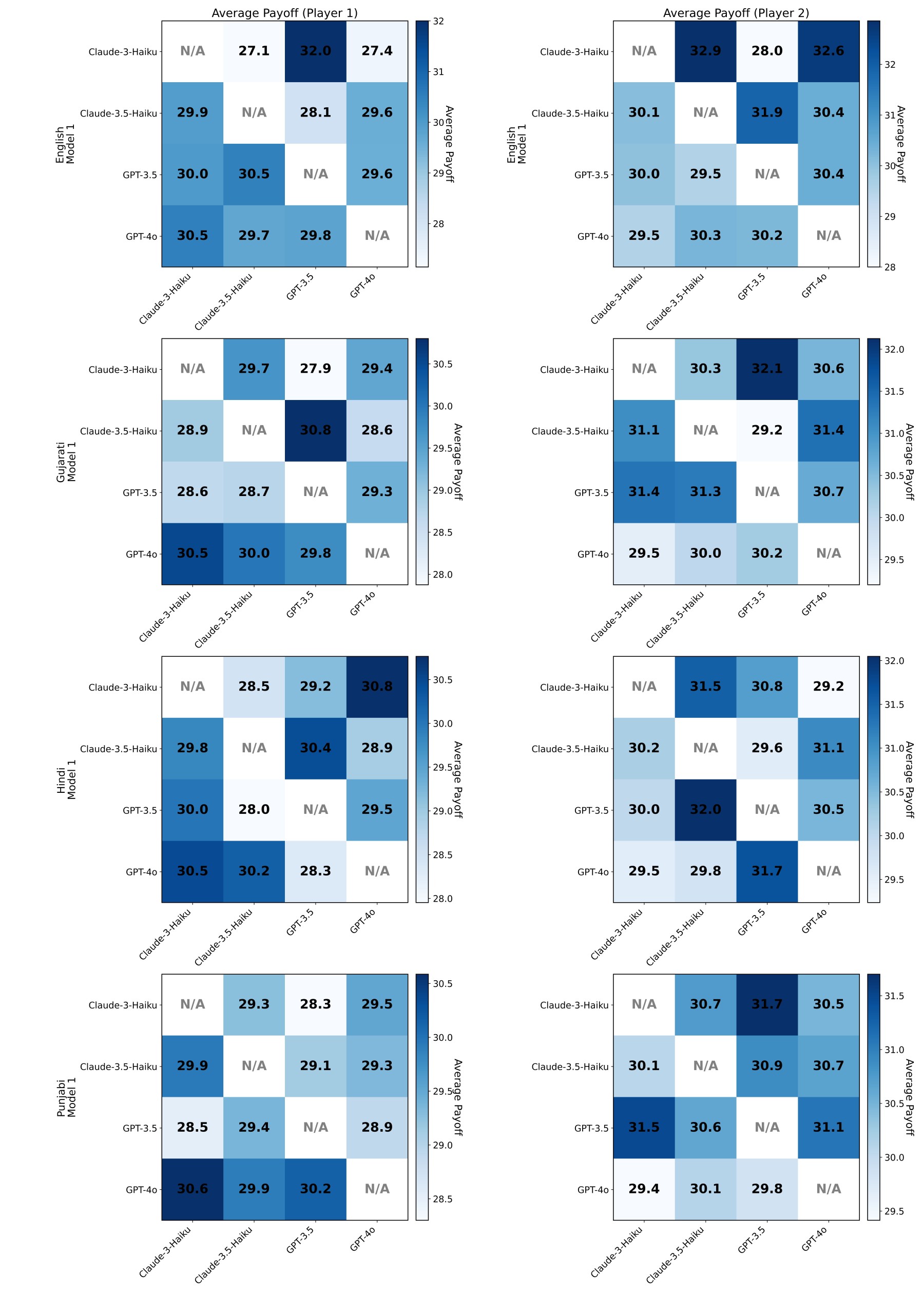}
\end{center}
\captionof{figure}{Heatmaps for \textbf{Resource Exchange Game} comparing average payoffs for P1 and P2 for model combinations for all languages.}
\label{fig:trading_allheatmaps}
\clearpage

\section{Statistical Analysis}
\label{sec:statanaly}

\begin{table*}[h!]
\centering
\small
\setlength{\tabcolsep}{4pt}
\renewcommand{\arraystretch}{1.2}

\begin{tabular}{llccccc}
\toprule
\textbf{Metric} & \textbf{Comparison} & \textbf{Mean Diff} & \textbf{U / z} & \textbf{p} & \textbf{p\_corr} & \textbf{Sig} \\
\midrule

\multicolumn{7}{l}{\textbf{P1 Payoff (H=10.69, p=0.0135, *)}} \\
\midrule
 & Baseline vs Gujarati & 3.45 & 70828.5 & 0.026187 & 0.052375 & ns \\
 & Baseline vs Hindi & 2.66 & 68838.0 & 0.135475 & 0.203213 & ns \\
 & Baseline vs Punjabi & -1.21 & 61928.5 & 0.290632 & 0.348759 & ns \\
 & Gujarati vs Hindi & -0.79 & 62568.5 & 0.411617 & 0.411617 & ns \\
 & Gujarati vs Punjabi & -4.66 & 57412.0 & 0.006761 & 0.040569 & * \\
 & Hindi vs Punjabi & -3.87 & 58609.0 & 0.023011 & 0.052375 & ns \\

\midrule
\multicolumn{7}{l}{\textbf{P2 Payoff (H=5.42, p=0.143, ns)}} \\
\midrule
 & Baseline vs Gujarati & 0.99 & 64262.5 & 0.842791 & 0.842791 & ns \\
 & Baseline vs Hindi & -0.72 & 63148.0 & 0.541194 & 0.811790 & ns \\
 & Baseline vs Punjabi & 2.87 & 69765.5 & 0.067414 & 0.202242 & ns \\
 & Gujarati vs Hindi & -1.71 & 63883.0 & 0.735474 & 0.842791 & ns \\
 & Gujarati vs Punjabi & 1.88 & 68884.0 & 0.133563 & 0.267125 & ns \\
 & Hindi vs Punjabi & 3.59 & 70756.0 & 0.028565 & 0.171392 & ns \\

\midrule
\multicolumn{7}{l}{\textbf{Initial Offer (H=49.27, p=1.14e-10, ***)}} \\
\midrule
 & Baseline vs Gujarati & 0.11 & 62883.0 & 0.735124 & 0.735124 & ns \\
 & Baseline vs Hindi & -1.77 & 60194.5 & 0.134966 & 0.202448 & ns \\
 & Baseline vs Punjabi & 5.36 & 77324.0 & 0.000000 & 0.000000 & *** \\
 & Gujarati vs Hindi & -1.88 & 60569.5 & 0.287276 & 0.344731 & ns \\
 & Gujarati vs Punjabi & 5.25 & 75896.5 & 0.000000 & 0.000001 & *** \\
 & Hindi vs Punjabi & 7.13 & 79118.5 & 0.000000 & 0.000000 & *** \\

\midrule
\multicolumn{7}{l}{\textbf{Total Turns (H=45.40, p=7.60e-10, ***)}} \\
\midrule
 & Baseline vs Gujarati & -0.55 & 50565.0 & 0.000000 & 0.000000 & *** \\
 & Baseline vs Hindi & -0.49 & 51253.0 & 0.000000 & 0.000000 & *** \\
 & Baseline vs Punjabi & -0.59 & 50900.5 & 0.000000 & 0.000000 & *** \\
 & Gujarati vs Hindi & 0.06 & 65987.5 & 0.649083 & 0.778899 & ns \\
 & Gujarati vs Punjabi & -0.04 & 64483.5 & 0.903086 & 0.903086 & ns \\
 & Hindi vs Punjabi & -0.10 & 63510.0 & 0.618610 & 0.778899 & ns \\

\midrule
\multicolumn{7}{l}{\textbf{Acceptance Rate ($\chi^2$=13.77, p=0.0032, **)}} \\
\midrule
 & Baseline vs Gujarati & 0.086 & 3.656 & 0.000256 & 0.001536 & ** \\
 & Baseline vs Hindi & 0.047 & 2.181 & 0.029196 & 0.058392 & ns \\
 & Baseline vs Punjabi & 0.064 & 2.840 & 0.004515 & 0.013545 & * \\
 & Gujarati vs Hindi & -0.039 & -1.522 & 0.128122 & 0.192183 & ns \\
 & Gujarati vs Punjabi & -0.022 & -0.848 & 0.396380 & 0.475656 & ns \\
 & Hindi vs Punjabi & 0.017 & 0.676 & 0.498962 & 0.498962 & ns \\

\bottomrule
\end{tabular}

\caption{Statistical comparison across languages for the \textbf{Ultimatum} Game. Global tests (Kruskal-Wallis or Chi-square) are reported per metric. Pairwise comparisons use Mann-Whitney U tests (or proportion z-tests for acceptance rate) with Benjamini-Hochberg correction.}
\label{tab:ultimatum_stats}
\end{table*}

\begin{table*}[h!]
\centering
\small
\setlength{\tabcolsep}{4pt}
\renewcommand{\arraystretch}{1.2}

\begin{tabular}{llcccccc}
\toprule
\textbf{Metric} & \textbf{Combination} & \textbf{Mean Diff} & \textbf{U} & \textbf{p} & \textbf{p\_corr} & \textbf{Sig} \\
\midrule

\multicolumn{7}{l}{\textbf{Seller Advantage (H=1.52, p=0.678, ns)}} \\
\midrule
 & English vs Gujarati & -0.81 & 56867.5 & 0.3373 & 0.8758 & ns \\
 & English vs Hindi & -1.10 & 53857.5 & 0.8758 & 0.8758 & ns \\
 & English vs Punjabi & -1.05 & 52309.0 & 0.8430 & 0.8758 & ns \\
 & Gujarati vs Hindi & -0.29 & 48463.0 & 0.2300 & 0.8758 & ns \\
 & Gujarati vs Punjabi & -0.24 & 47596.0 & 0.5408 & 0.8758 & ns \\
 & Hindi vs Punjabi & 0.05 & 49752.5 & 0.6166 & 0.8758 & ns \\

\midrule
\multicolumn{7}{l}{\textbf{Buyer Advantage (H=1.52, p=0.678, ns)}} \\
\midrule
 & English vs Gujarati & 0.81 & 52272.5 & 0.3373 & 0.8758 & ns \\
 & English vs Hindi & 1.10 & 54602.5 & 0.8758 & 0.8758 & ns \\
 & English vs Punjabi & 1.05 & 51391.0 & 0.8430 & 0.8758 & ns \\
 & Gujarati vs Hindi & 0.29 & 53936.0 & 0.2300 & 0.8758 & ns \\
 & Gujarati vs Punjabi & 0.24 & 50309.0 & 0.5408 & 0.8758 & ns \\
 & Hindi vs Punjabi & -0.05 & 47542.5 & 0.6166 & 0.8758 & ns \\

\midrule
\multicolumn{7}{l}{\textbf{Negotiation Rounds (H=14.33, p=0.0025, **)}} \\
\midrule
 & English vs Gujarati & -0.22 & 52309.5 & 0.0481 & 0.0962 & ns \\
 & English vs Hindi & 0.05 & 52706.5 & 0.1704 & 0.2045 & ns \\
 & English vs Punjabi & -0.33 & 45490.0 & 0.0002 & 0.0012 & ** \\
 & Gujarati vs Hindi & 0.28 & 53571.0 & 0.4272 & 0.4272 & ns \\
 & Gujarati vs Punjabi & -0.11 & 46711.0 & 0.1225 & 0.1838 & ns \\
 & Hindi vs Punjabi & -0.38 & 43977.5 & 0.0156 & 0.0468 & * \\

\midrule
\multicolumn{7}{l}{\textbf{Acceptance Rate}} \\
\midrule
\multicolumn{7}{l}{English: 340/350 (97.1\%)} \\
\multicolumn{7}{l}{Gujarati: 321/325 (98.8\%)} \\
\multicolumn{7}{l}{Hindi: 319/319 (100.0\%)} \\
\multicolumn{7}{l}{Punjabi: 306/308 (99.4\%)} \\
\multicolumn{7}{l}{Chi-square test skipped (degenerate case)} \\

\bottomrule
\end{tabular}

\caption{Statistical comparison across languages for the \textbf{BuySell} Game. Global tests (Kruskal-Wallis or Chi-square) are shown alongside each metric. Pairwise comparisons use Mann-Whitney U tests (or proportion tests) with Benjamini-Hochberg correction.}
\label{tab:buysell_stats}
\end{table*}

\begin{table*}[h!]
\centering
\small
\setlength{\tabcolsep}{4pt}
\renewcommand{\arraystretch}{1.2}

\begin{tabular}{llccc}
\toprule
\textbf{Metric} & \textbf{Comparison} & \textbf{p} & \textbf{p\_corr} & \textbf{Sig} \\
\midrule

\multicolumn{5}{l}{\textbf{Trade Volume (H=30.42, p=1.13e-06, ***)}} \\
\midrule
 & English vs Gujarati & 0.0001 & 0.0001 & *** \\
 & English vs Hindi & 0.0000 & 0.0000 & *** \\
 & English vs Punjabi & 0.0000 & 0.0000 & *** \\
 & Gujarati vs Hindi & 0.5743 & 0.8079 & ns \\
 & Gujarati vs Punjabi & 0.7728 & 0.8079 & ns \\
 & Hindi vs Punjabi & 0.8079 & 0.8079 & ns \\

\midrule
\multicolumn{5}{l}{\textbf{P1 Payoff (H=1.09, p=0.780, ns)}} \\
\midrule
 & English vs Gujarati & 0.5929 & 0.8249 & ns \\
 & English vs Hindi & 0.5954 & 0.8249 & ns \\
 & English vs Punjabi & 0.9096 & 0.9096 & ns \\
 & Gujarati vs Hindi & 0.3222 & 0.8249 & ns \\
 & Gujarati vs Punjabi & 0.5075 & 0.8249 & ns \\
 & Hindi vs Punjabi & 0.6874 & 0.8249 & ns \\

\midrule
\multicolumn{5}{l}{\textbf{P2 Payoff (H=1.09, p=0.780, ns)}} \\
\midrule
 & English vs Gujarati & 0.5929 & 0.8249 & ns \\
 & English vs Hindi & 0.5954 & 0.8249 & ns \\
 & English vs Punjabi & 0.9096 & 0.9096 & ns \\
 & Gujarati vs Hindi & 0.3222 & 0.8249 & ns \\
 & Gujarati vs Punjabi & 0.5075 & 0.8249 & ns \\
 & Hindi vs Punjabi & 0.6874 & 0.8249 & ns \\

\midrule
\multicolumn{5}{l}{\textbf{Negotiation Rounds (H=4.95, p=0.175, ns)}} \\
\midrule
 & English vs Gujarati & 0.0590 & 0.2381 & ns \\
 & English vs Hindi & 0.0794 & 0.2381 & ns \\
 & English vs Punjabi & 0.4542 & 0.5450 & ns \\
 & Gujarati vs Hindi & 0.9426 & 0.9426 & ns \\
 & Gujarati vs Punjabi & 0.2148 & 0.3911 & ns \\
 & Hindi vs Punjabi & 0.2608 & 0.3911 & ns \\

\midrule
\multicolumn{5}{l}{\textbf{Win Rate (P1) ($\chi^2$=2.26, p=0.520, ns)}} \\
\midrule
 & English vs Gujarati & 0.8521 & 0.9947 & ns \\
 & English vs Hindi & 0.2746 & 0.5492 & ns \\
 & English vs Punjabi & 0.9947 & 0.9947 & ns \\
 & Gujarati vs Hindi & 0.1838 & 0.5492 & ns \\
 & Gujarati vs Punjabi & 0.8418 & 0.9947 & ns \\
 & Hindi vs Punjabi & 0.2607 & 0.5492 & ns \\

\midrule
\multicolumn{5}{l}{\textbf{Acceptance Rate ($\chi^2$ test: not applicable, degenerate case)}} \\

\bottomrule
\end{tabular}

\caption{Statistical comparison across languages for the \textbf{Resource Exchange} Game. Global tests (Kruskal-Wallis or Chi-square) are shown alongside each metric. Pairwise comparisons use Mann-Whitney U tests (or proportion tests) with Benjamini-Hochberg correction.}
\label{tab:trading_stats}
\end{table*}

\section{Language Compliance}
\label{sec:langcompliance}

\begin{table*}[h!]
\centering
\small
\setlength{\tabcolsep}{4pt}
\begin{tabular}{llcc}
\toprule
\textbf{Game} & \textbf{Lang.} & \textbf{Rate} & \textbf{Conf.} \\
\midrule
Ultimatum & Gujarati & 97.02\% & 0.999 \\
          & Hindi    & 92.18\% & 0.966 \\
          & Punjabi  & 94.84\% & 0.998 \\
\midrule
Buy-Sell  & Gujarati & 93.22\% & 0.998 \\
          & Hindi    & 94.47\% & 0.981 \\
          & Punjabi  & 89.65\% & 0.997 \\
\midrule
Res.      & Gujarati & 97.85\% & 0.999 \\
Exchange  & Hindi    & 94.97\% & 0.971 \\
          & Punjabi  & 90.70\% & 0.997 \\
\bottomrule
\end{tabular}
\caption{Language adherence rates and average confidence 
scores per game and language.}
\label{tab:lang_adherence}
\end{table*}

\section{Prompt Sensitivity Ablation Details}
\label{sec:promptablation}

\subsection{Prompt Ablation 1}

\begin{figure*}[h]
    \centering
    \includegraphics[width=0.9\linewidth]{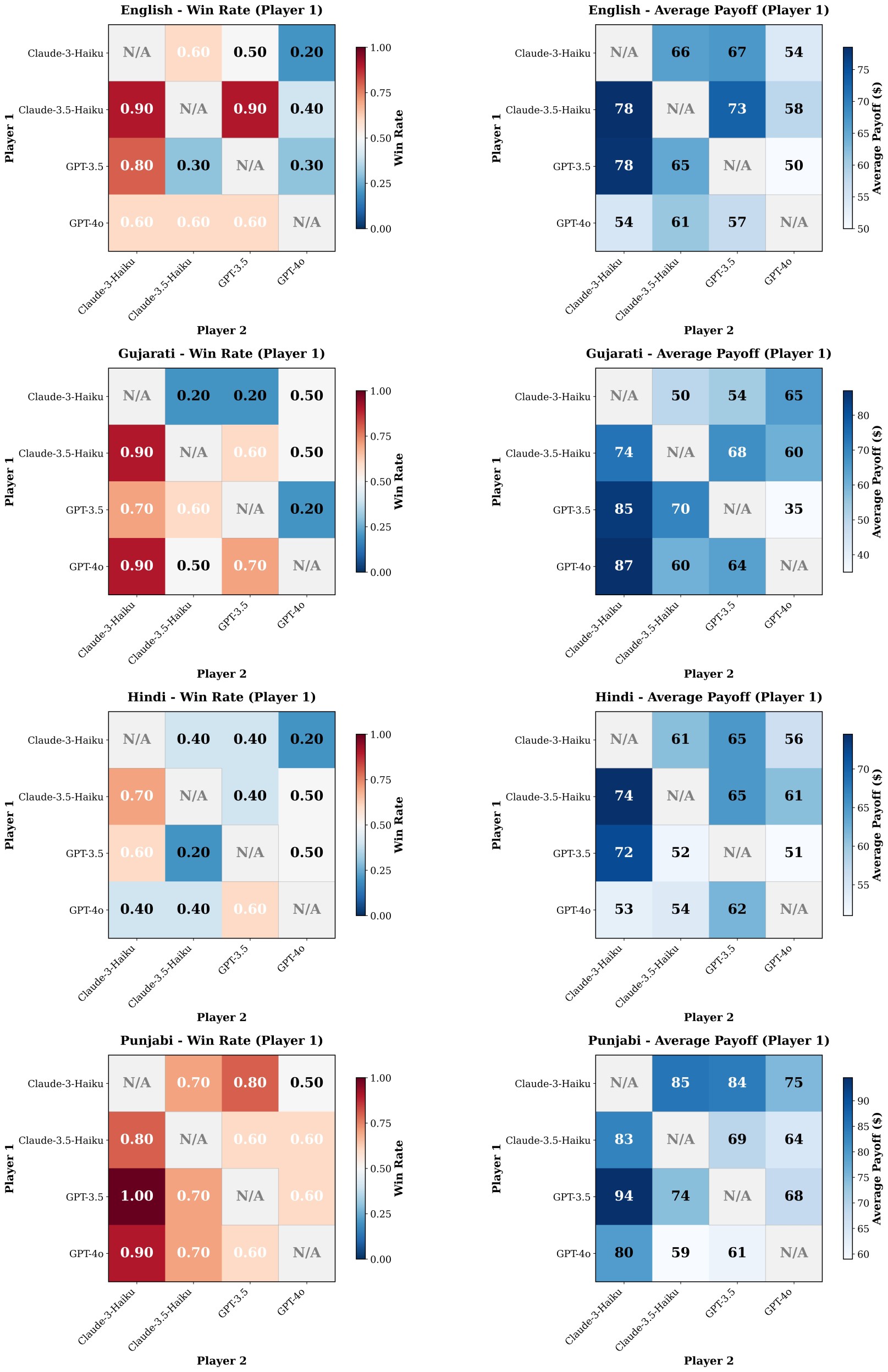}
    \caption{Heatmaps for \textbf{Ultimatum Game} prompt sensitivity ablation 1 comparing model combination outcomes across all languages.}
    \label{fig:ulti_heat_prompt1}
\end{figure*}

See Figure~\ref{fig:ulti_heat_prompt1} for model comparison heatmaps, Table~\ref{tab:ultimatum_summary_prompt1} for the data summarized per language and Table~\ref{tab:ultimatum_stats_prompt1} for the statistical analysis.

\begin{table*}[t]
\centering
\small
\setlength{\tabcolsep}{6pt}
\begin{tabular}{llccc}
\toprule
\textbf{Metric} & \textbf{Comparison} & \textbf{p} & \textbf{p\_corr} & \textbf{Sig} \\
\midrule

\multicolumn{5}{l}{\textbf{Player 1 Payoff (H=26.252878, p=8.4426e-06, ***)}} \\
\midrule
 & English vs Gujarati & 0.581387 & 0.581387 & ns \\
 & English vs Hindi & 0.188140 & 0.225768 & ns \\
 & English vs Punjabi & 0.000028 & 0.000083 & *** \\
 & Gujarati vs Hindi & 0.145061 & 0.217591 & ns \\
 & Gujarati vs Punjabi & 0.003272 & 0.006545 & ** \\
 & Hindi vs Punjabi & 0.000002 & 0.000014 & *** \\

\midrule
\multicolumn{5}{l}{\textbf{Player 2 Payoff (H=31.897117, p=5.5014e-07, ***)}} \\
\midrule
 & English vs Gujarati & 0.058127 & 0.069752 & ns \\
 & English vs Hindi & 0.278217 & 0.278217 & ns \\
 & English vs Punjabi & 0.000003 & 0.000008 & *** \\
 & Gujarati vs Hindi & 0.010207 & 0.020414 & * \\
 & Gujarati vs Punjabi & 0.022788 & 0.034182 & * \\
 & Hindi vs Punjabi & 0.000000 & 0.000003 & *** \\

\midrule
\multicolumn{5}{l}{\textbf{Initial Offer (H=7.791326, p=5.0527e-02, ns)}} \\
\midrule
 & English vs Gujarati & 0.555858 & 0.667030 & ns \\
 & English vs Hindi & 0.319284 & 0.478926 & ns \\
 & English vs Punjabi & 0.075029 & 0.150059 & ns \\
 & Gujarati vs Hindi & 0.709829 & 0.709829 & ns \\
 & Gujarati vs Punjabi & 0.036438 & 0.109315 & ns \\
 & Hindi vs Punjabi & 0.011628 & 0.069769 & ns \\

\midrule
\multicolumn{5}{l}{\textbf{Total Turns (H=1.805516, p=6.1374e-01, ns)}} \\
\midrule
 & English vs Gujarati & 0.475385 & 0.713077 & ns \\
 & English vs Hindi & 0.636916 & 0.764299 & ns \\
 & English vs Punjabi & 0.406831 & 0.713077 & ns \\
 & Gujarati vs Hindi & 0.348166 & 0.713077 & ns \\
 & Gujarati vs Punjabi & 0.885536 & 0.885536 & ns \\
 & Hindi vs Punjabi & 0.245411 & 0.713077 & ns \\

\midrule
\multicolumn{5}{l}{\textbf{Acceptance Rate ($\chi^2$=26.005063, p=9.5142e-06, ***)}} \\
\midrule
 & English vs Gujarati & 0.000862 & 0.001842 & ** \\
 & English vs Hindi & 0.124783 & 0.140297 & ns \\
 & English vs Punjabi & 0.000003 & 0.000018 & *** \\
 & Gujarati vs Hindi & 0.059214 & 0.088821 & ns \\
 & Gujarati vs Punjabi & 0.140297 & 0.140297 & ns \\
 & Hindi vs Punjabi & 0.000921 & 0.001842 & ** \\

\bottomrule
\end{tabular}

\caption{Statistical comparison across languages for prompt ablation 1 of the \textbf{Ultimatum Game}. Global tests (Kruskal-Wallis or Chi-square) are shown alongside each metric. Pairwise comparisons use Mann-Whitney U tests (or proportion tests) with Benjamini-Hochberg correction.}
\label{tab:ultimatum_stats_prompt1}
\end{table*}

\begin{table*}[htbp]
\centering
\resizebox{\textwidth}{!}{%
\small
\begin{tabular}{lcccccc}
\toprule
Language & Acceptance Rate & Initial Offer & P1 Payoff & P2 Payoff & P1 Win Rate & Conversation Rounds \\
\midrule
English & \textbf{93.33\% $\pm$ 24.94\%} & 41.43 $\pm$ 11.69 & 63.50 $\pm$ 19.16 & 34.83 $\pm$ 17.86 & 55.83\% $\pm$ 49.66\% & 2.63 $\pm$ 1.17 \\
Gujarati & 78.33\% $\pm$ 41.20\% & 41.65 $\pm$ 14.43 & 64.25 $\pm$ 28.35 & 29.08 $\pm$ 23.86 & 54.17\% $\pm$ 49.83\% & \textbf{2.82 $\pm$ 1.55} \\
Hindi & 87.50\% $\pm$ 33.07\% & \textbf{42.69 $\pm$ 13.79} & 60.54 $\pm$ 23.25 & \textbf{36.96 $\pm$ 21.95} & 44.17\% $\pm$ 49.66\% & 2.71 $\pm$ 1.40 \\
Punjabi & 70.00\% $\pm$ 45.83\% & 37.36 $\pm$ 16.11 & \textbf{74.71 $\pm$ 25.75} & 21.96 $\pm$ 22.07 & \textbf{70.83\% $\pm$ 45.45\%} & 2.87 $\pm$ 1.46 \\
\bottomrule
\end{tabular}
}
\caption{Metrics for prompt ablation 1 of the \textbf{Ultimatum Game} aggregated across all model combinations (mean $\pm$ std).}
\label{tab:ultimatum_summary_prompt1}
\end{table*}

\subsection{Prompt Ablation 2}

\begin{figure*}[h]
    \centering
    \includegraphics[width=0.9\linewidth]{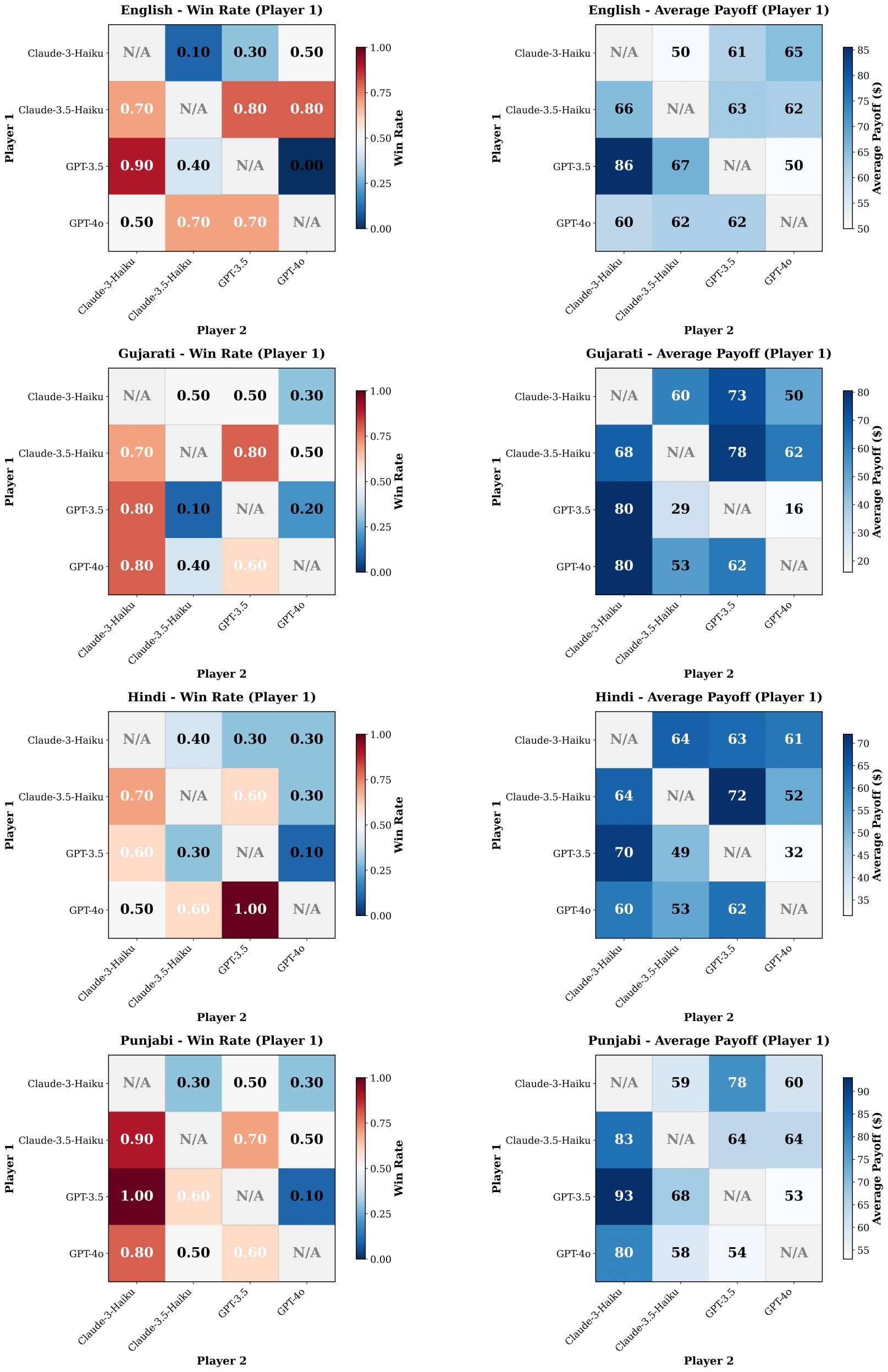}
    \caption{Heatmaps for \textbf{Ultimatum Game} prompt sensitivity ablation 2 comparing model combination outcomes across all languages.}
    \label{fig:ulti_heat_prompt2}
\end{figure*}

See Figure~\ref{fig:ulti_heat_prompt2} for model comparison heatmaps, Table~\ref{tab:ultimatum_summary_prompt2} for the data summarized per language and Table~\ref{tab:ultimatum_stats_prompt2} for the statistical analysis.

\begin{table*}[t]
\centering
\small
\setlength{\tabcolsep}{6pt}
\begin{tabular}{llccc}
\toprule
\textbf{Metric} & \textbf{Comparison} & \textbf{p} & \textbf{p\_corr} & \textbf{Sig} \\
\midrule

\multicolumn{5}{l}{\textbf{Player 1 Payoff (H=6.429618, p=9.2480e-02, ns)}} \\
\midrule
 & English vs Gujarati & 0.791180 & 0.791180 & ns \\
 & English vs Hindi & 0.188787 & 0.283180 & ns \\
 & English vs Punjabi & 0.146649 & 0.283180 & ns \\
 & Gujarati vs Hindi & 0.484535 & 0.581443 & ns \\
 & Gujarati vs Punjabi & 0.093983 & 0.281948 & ns \\
 & Hindi vs Punjabi & 0.016924 & 0.101542 & ns \\

\midrule
\multicolumn{5}{l}{\textbf{Player 2 Payoff (H=13.150719, p=4.3217e-03, **)}} \\
\midrule
 & English vs Gujarati & 0.018953 & 0.037907 & * \\
 & English vs Hindi & 0.276479 & 0.331774 & ns \\
 & English vs Punjabi & 0.056495 & 0.084743 & ns \\
 & Gujarati vs Hindi & 0.002647 & 0.015882 & * \\
 & Gujarati vs Punjabi & 0.799788 & 0.799788 & ns \\
 & Hindi vs Punjabi & 0.008657 & 0.025970 & * \\

\midrule
\multicolumn{5}{l}{\textbf{Initial Offer (H=12.815051, p=5.0541e-03, **)}} \\
\midrule
 & English vs Gujarati & 0.468504 & 0.468504 & ns \\
 & English vs Hindi & 0.253647 & 0.304377 & ns \\
 & English vs Punjabi & 0.009133 & 0.027400 & * \\
 & Gujarati vs Hindi & 0.099666 & 0.149499 & ns \\
 & Gujarati vs Punjabi & 0.084277 & 0.149499 & ns \\
 & Hindi vs Punjabi & 0.001100 & 0.006599 & ** \\

\midrule
\multicolumn{5}{l}{\textbf{Total Turns (H=10.875555, p=1.2418e-02, *)}} \\
\midrule
 & English vs Gujarati & 0.002190 & 0.013143 & * \\
 & English vs Hindi & 0.277313 & 0.415969 & ns \\
 & English vs Punjabi & 0.478605 & 0.574326 & ns \\
 & Gujarati vs Hindi & 0.036887 & 0.073773 & ns \\
 & Gujarati vs Punjabi & 0.030010 & 0.073773 & ns \\
 & Hindi vs Punjabi & 0.767648 & 0.767648 & ns \\

\midrule
\multicolumn{5}{l}{\textbf{Acceptance Rate ($\chi^2$=24.532418, p=1.9337e-05, ***)}} \\
\midrule
 & English vs Gujarati & 0.000032 & 0.000194 & *** \\
 & English vs Hindi & 0.253369 & 0.304043 & ns \\
 & English vs Punjabi & 0.000174 & 0.000522 & *** \\
 & Gujarati vs Hindi & 0.001677 & 0.003354 & ** \\
 & Gujarati vs Punjabi & 0.656486 & 0.656486 & ns \\
 & Hindi vs Punjabi & 0.006565 & 0.009848 & ** \\

\bottomrule
\end{tabular}

\caption{Statistical comparison across languages for prompt ablation 2 of the \textbf{Ultimatum Game}. Global tests (Kruskal-Wallis or Chi-square) are shown alongside each metric. Pairwise comparisons use Mann-Whitney U tests (or proportion tests) with Benjamini-Hochberg correction.}
\label{tab:ultimatum_stats_prompt2}
\end{table*}

\begin{table*}[htbp]
\centering
\resizebox{\textwidth}{!}{%
\small
\begin{tabular}{lcccccc}
\toprule
Language & Acceptance Rate & Initial Offer & P1 Payoff & P2 Payoff & P1 Win Rate & Conversation Rounds \\
\midrule
English & \textbf{93.33\% $\pm$ 24.94\%} & 43.04 $\pm$ 10.61 & 62.79 $\pm$ 18.12 & 36.38 $\pm$ 17.50 & 53.33\% $\pm$ 49.89\% & 2.51 $\pm$ 1.03 \\
Gujarati & 73.33\% $\pm$ 44.22\% & 41.56 $\pm$ 12.41 & 59.34 $\pm$ 30.32 & 29.82 $\pm$ 24.49 & 51.67\% $\pm$ 49.97\% & \textbf{2.93 $\pm$ 1.61} \\
Hindi & 89.17\% $\pm$ 31.08\% & \textbf{46.38 $\pm$ 17.14} & 58.52 $\pm$ 24.04 & \textbf{39.82 $\pm$ 23.38} & 47.50\% $\pm$ 49.94\% & 2.66 $\pm$ 1.19 \\
Punjabi & 75.83\% $\pm$ 42.81\% & 39.13 $\pm$ 15.79 & \textbf{67.83 $\pm$ 26.12} & 29.67 $\pm$ 24.23 & \textbf{56.67\% $\pm$ 49.55\%} & 2.68 $\pm$ 1.29 \\
\bottomrule
\end{tabular}
}
\caption{Metrics for prompt ablation 2 of the \textbf{Ultimatum Game} aggregated across all model combinations (mean $\pm$ std).}
\label{tab:ultimatum_summary_prompt2}
\end{table*}

\subsection{Native Language Prompt Ablation}

\begin{figure*}[h]
    \centering
    \includegraphics[width=0.9\linewidth]{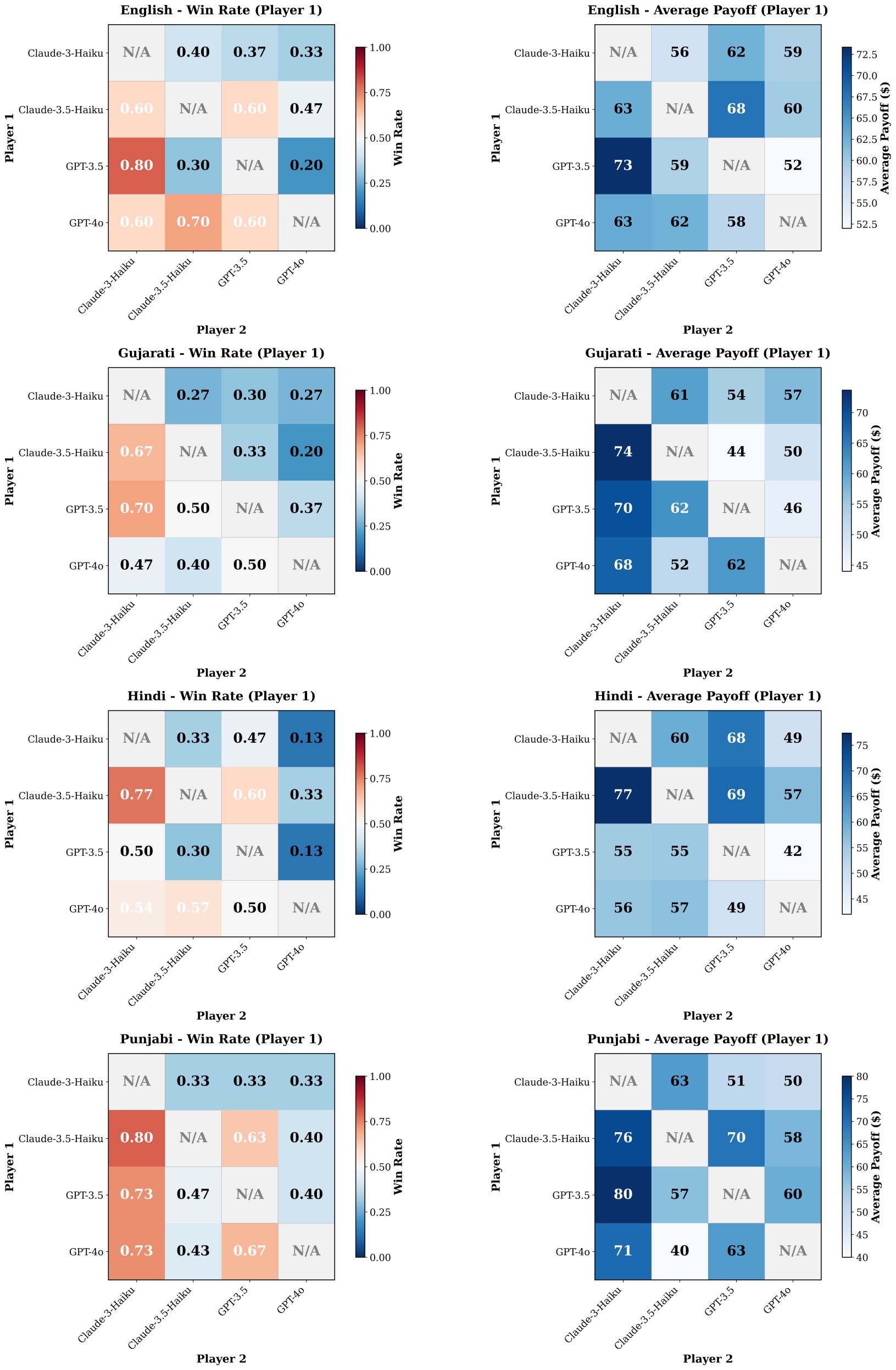}
    \caption{Heatmaps for \textbf{Ultimatum Game} with prompt in native language comparing model combination outcomes across all languages.}
    \label{fig:ulti_heat_native}
\end{figure*}

See Figure~\ref{fig:ulti_heat_native} for model comparison heatmaps, Table~\ref{tab:ultimatum_summary_promptnative} for the data summarized per language and Table~\ref{tab:ultimatum_stats_promptnative} for the statistical analysis.

\begin{table*}[t]
\centering
\small
\setlength{\tabcolsep}{6pt}
\begin{tabular}{llccc}
\toprule
\textbf{Metric} & \textbf{Comparison} & \textbf{p} & \textbf{p\_corr} & \textbf{Sig} \\
\midrule

\multicolumn{5}{l}{\textbf{Player 1 Payoff (H=7.105005, p=6.8625e-02, ns)}} \\
\midrule
 & English vs Gujarati & 0.219148 & 0.269483 & ns \\
 & English vs Hindi & 0.146722 & 0.269483 & ns \\
 & English vs Punjabi & 0.224569 & 0.269483 & ns \\
 & Gujarati vs Hindi & 0.838844 & 0.838844 & ns \\
 & Gujarati vs Punjabi & 0.054615 & 0.163845 & ns \\
 & Hindi vs Punjabi & 0.019696 & 0.118175 & ns \\

\midrule
\multicolumn{5}{l}{\textbf{Player 2 Payoff (H=40.474844, p=8.4506e-09, ***)}} \\
\midrule
 & English vs Gujarati & 0.005886 & 0.008829 & ** \\
 & English vs Hindi & 0.887990 & 0.887990 & ns \\
 & English vs Punjabi & 0.000000 & 0.000000 & *** \\
 & Gujarati vs Hindi & 0.004693 & 0.008829 & ** \\
 & Gujarati vs Punjabi & 0.017205 & 0.020646 & * \\
 & Hindi vs Punjabi & 0.000000 & 0.000000 & *** \\

\midrule
\multicolumn{5}{l}{\textbf{Initial Offer (H=17.798184, p=4.8408e-04, ***)}} \\
\midrule
 & English vs Gujarati & 0.020541 & 0.030812 & * \\
 & English vs Hindi & 0.006024 & 0.012048 & * \\
 & English vs Punjabi & 0.257185 & 0.308622 & ns \\
 & Gujarati vs Hindi & 0.550578 & 0.550578 & ns \\
 & Gujarati vs Punjabi & 0.002303 & 0.006910 & ** \\
 & Hindi vs Punjabi & 0.000654 & 0.003923 & ** \\

\midrule
\multicolumn{5}{l}{\textbf{Total Turns (H=22.020259, p=6.4601e-05, ***)}} \\
\midrule
 & English vs Gujarati & 0.000008 & 0.000049 & *** \\
 & English vs Hindi & 0.000395 & 0.001184 & ** \\
 & English vs Punjabi & 0.002921 & 0.005843 & ** \\
 & Gujarati vs Hindi & 0.219151 & 0.328726 & ns \\
 & Gujarati vs Punjabi & 0.278274 & 0.333929 & ns \\
 & Hindi vs Punjabi & 0.987905 & 0.987905 & ns \\

\midrule
\multicolumn{5}{l}{\textbf{Acceptance Rate ($\chi^2$=67.810503, p=1.2560e-14, ***)}} \\
\midrule
 & English vs Gujarati & 0.000000 & 0.000000 & *** \\
 & English vs Hindi & 0.000744 & 0.001116 & ** \\
 & English vs Punjabi & 0.000000 & 0.000000 & *** \\
 & Gujarati vs Hindi & 0.005235 & 0.006282 & ** \\
 & Gujarati vs Punjabi & 0.056032 & 0.056032 & ns \\
 & Hindi vs Punjabi & 0.000003 & 0.000006 & *** \\

\bottomrule
\end{tabular}

\caption{Statistical comparison across languages for the native language prompt ablation of the \textbf{Ultimatum Game}. Global tests (Kruskal-Wallis or Chi-square) are shown alongside each metric. Pairwise comparisons use Mann-Whitney U tests (or proportion tests) with Benjamini-Hochberg correction.}
\label{tab:ultimatum_stats_promptnative}
\end{table*}

\begin{table*}[htbp]
\centering
\resizebox{\textwidth}{!}{%
\small
\begin{tabular}{lcccccc}
\toprule
Language & Acceptance Rate & Initial Offer & P1 Payoff & P2 Payoff & P1 Win Rate & Conversation Rounds \\
\midrule
English & \textbf{91.94\% $\pm$ 27.22\%} & 42.75 $\pm$ 11.54 & 61.25 $\pm$ 20.05 & 37.08 $\pm$ 19.02 & 49.72\% $\pm$ 50.00\% & \textbf{2.75 $\pm$ 1.28} \\
Gujarati & 75.28\% $\pm$ 43.14\% & 44.55 $\pm$ 13.89 & 58.32 $\pm$ 29.66 & 31.40 $\pm$ 24.56 & 41.39\% $\pm$ 49.25\% & 2.39 $\pm$ 1.31 \\
Hindi & 83.71\% $\pm$ 36.93\% & \textbf{46.00 $\pm$ 18.27} & 57.92 $\pm$ 25.70 & \textbf{37.58 $\pm$ 23.85} & 42.98\% $\pm$ 49.50\% & 2.45 $\pm$ 1.17 \\
Punjabi & 68.89\% $\pm$ 46.29\% & 41.12 $\pm$ 16.12 & \textbf{61.51 $\pm$ 30.45} & 27.38 $\pm$ 23.40 & \textbf{52.22\% $\pm$ 49.95\%} & 2.58 $\pm$ 1.52 \\
\bottomrule
\end{tabular}
}
\caption{Metrics for the native language prompt ablation of the \textbf{Ultimatum Game} aggregated across all model combinations (mean $\pm$ std).}
\label{tab:ultimatum_summary_promptnative}
\end{table*}

\subsection{Comparing all Prompt Ablations}

\begin{figure*}[t]
    \centering
    \includegraphics[width=0.9\linewidth]{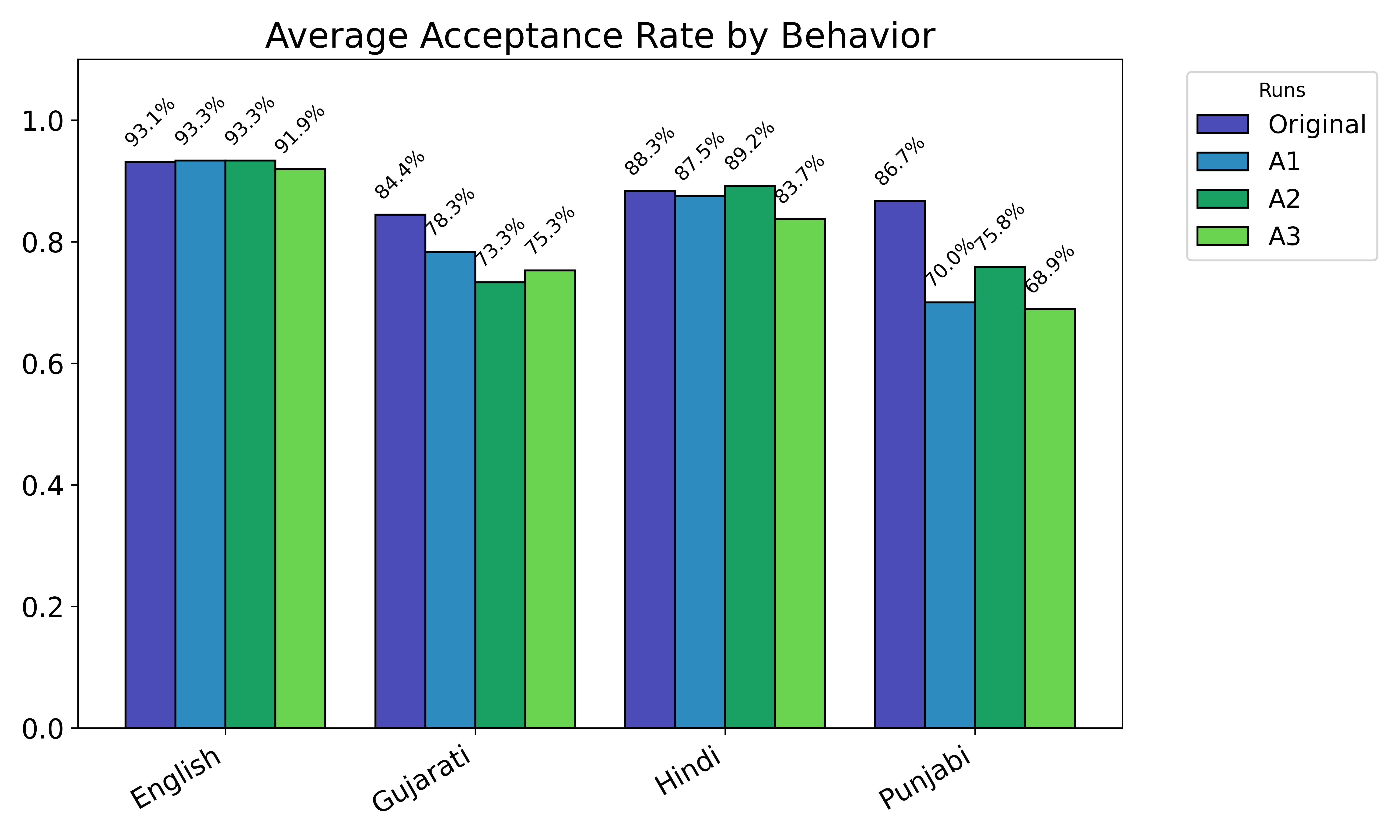}
    \caption{Acceptance Rate Comparison for \textbf{Ultimatum Game} comparing prompt ablation outcomes across all languages.}
    \label{fig:acceptratecompare}
\end{figure*}

\begin{figure*}[h]
    \centering
    \includegraphics[width=0.9\linewidth]{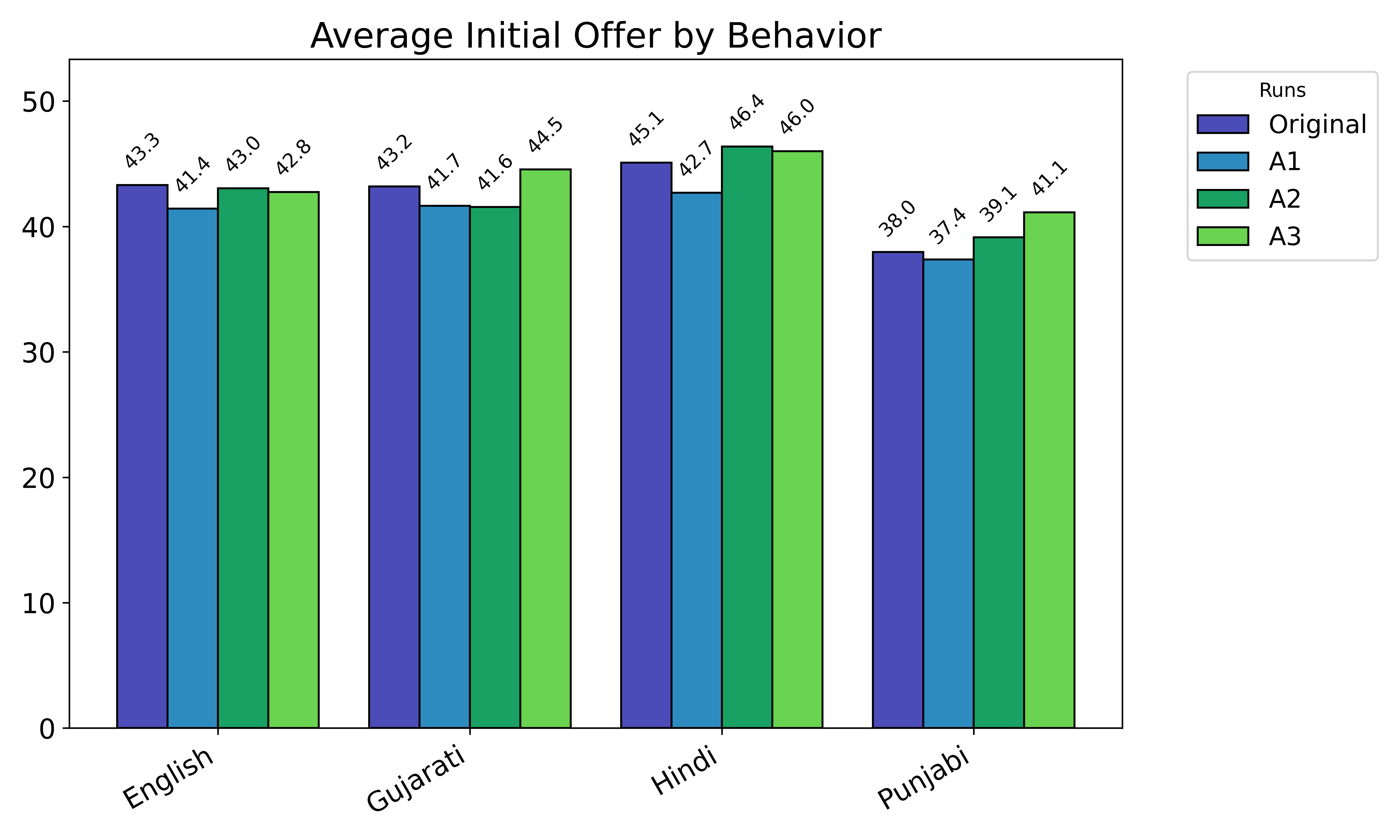}
    \caption{Initial Offer Comparison for \textbf{Ultimatum Game} comparing prompt ablation outcomes across all languages.}
    \label{fig:initialoffercompare}
\end{figure*}

\begin{figure*}[h]
    \centering
    \includegraphics[width=0.9\linewidth]{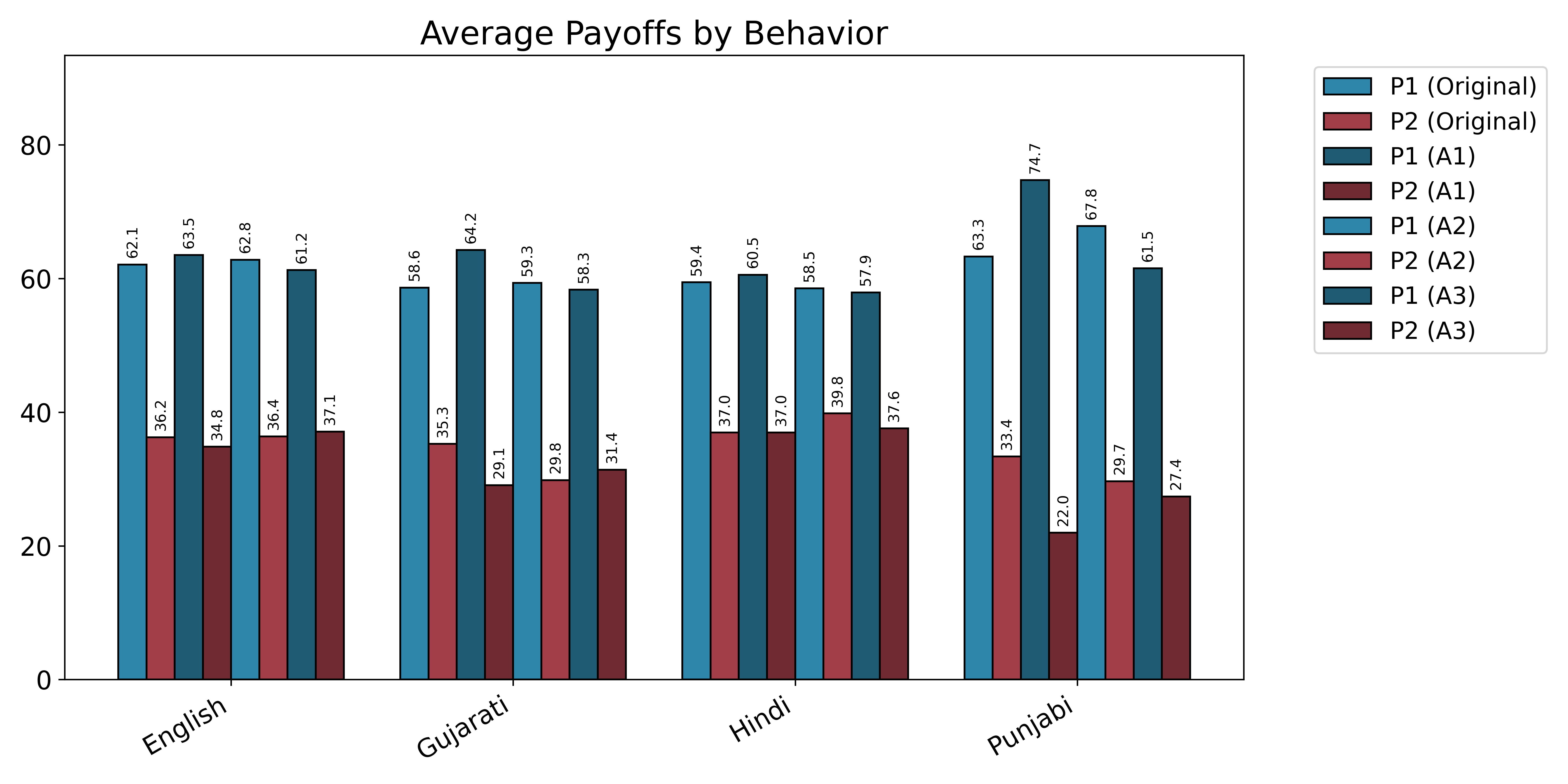}
    \caption{Payoffs Comparison for \textbf{Ultimatum Game} comparing prompt ablation outcomes across all languages.}
    \label{fig:payoffscompare}
\end{figure*}

\begin{figure*}[h]
    \centering
    \includegraphics[width=0.9\linewidth]{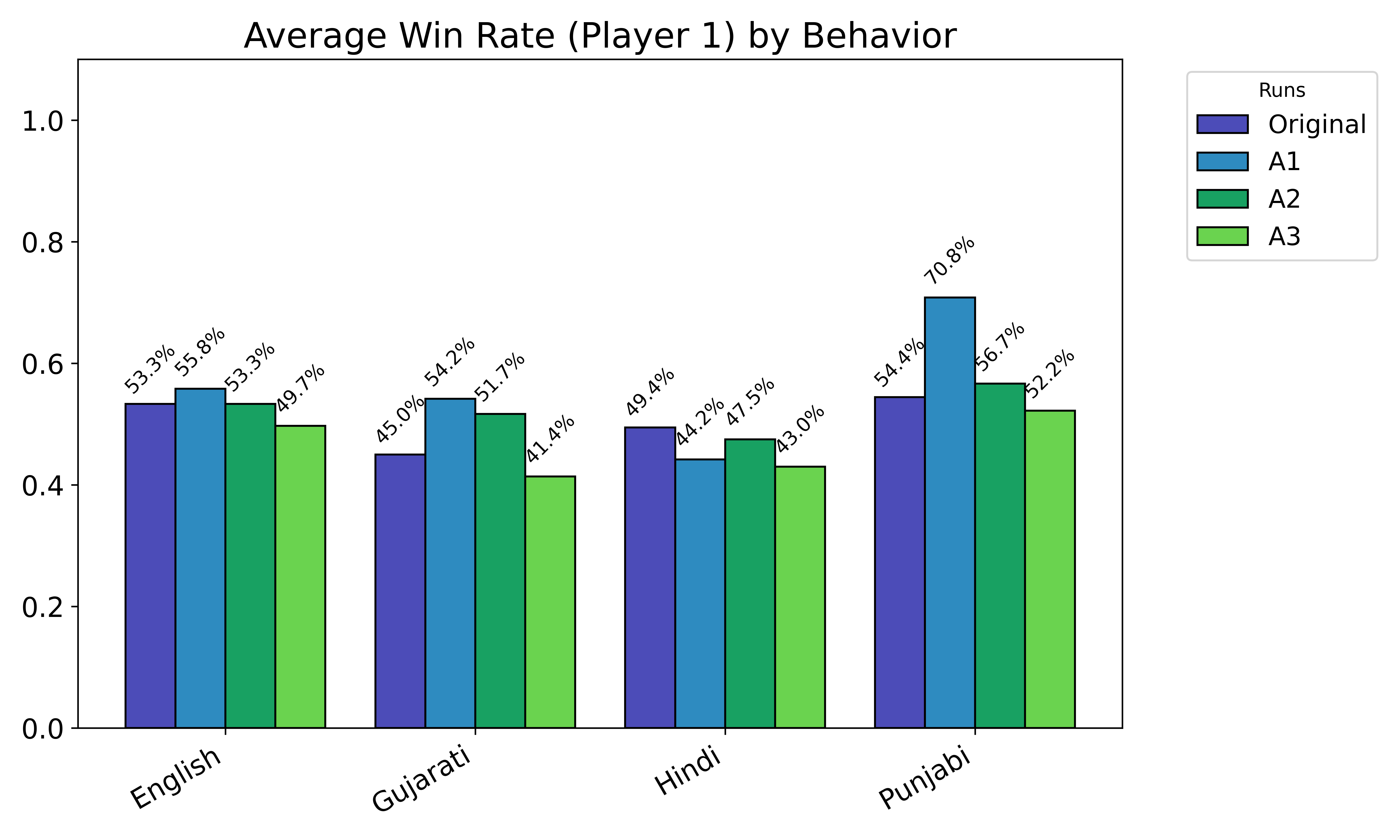}
    \caption{Payoffs Comparison for \textbf{Ultimatum Game} comparing prompt ablation outcomes across all languages.}
    \label{fig:winratecompare}
\end{figure*}

\end{document}